\documentclass{article}

\PassOptionsToPackage{numbers}{natbib}
\usepackage[main, final]{neurips_2026}
\usepackage[utf8]{inputenc}
\usepackage[T1]{fontenc}
\usepackage{url}
\usepackage{booktabs}
\usepackage{amsfonts}
\usepackage{amsmath}
\usepackage{amssymb}
\usepackage{amsthm}
\usepackage{mathtools}
\usepackage{nicefrac}
\usepackage{microtype}
\usepackage{xcolor}
\usepackage{wrapfig}
\usepackage[normalem]{ulem}

\usepackage{graphicx}
\usepackage{tikz}
\usetikzlibrary{arrows.meta, positioning, shapes.geometric, fit, calc, backgrounds}
\usepackage{subcaption}
\usepackage{enumitem}
\usepackage{flafter}
\usepackage[pagebackref=true]{hyperref} 

\usepackage{fontawesome5}

\usepackage{etoc}

\newsavebox{\celebatextbox}
\newsavebox{\celebacapbox}

\renewcommand*{\backref}[1]{} % Clear the default format
\renewcommand*{\backrefalt}[4]{%
  \ifcase #1 % Case for 0 citations
    (Not cited.)%
  \or % Case for 1 citation
    (Cited on page #4.)%
  \else % Case for 2+ citations
    (Cited on pages #4.)%
  \fi%
}

\newcommand{\R}{\mathbb{R}}
\newcommand{\E}{\mathbb{E}}
\newcommand{\X}{\mathcal{X}}

\newcommand{\Hc}{\mathcal{H}}
\DeclareMathOperator*{\argmin}{arg\,min}
\DeclareMathOperator{\tr}{tr}

\DeclareMathOperator{\Cov}{Cov}
\DeclareMathOperator{\rank}{rank}

\newcommand{\norm}[1]{\left\lVert #1 \right\rVert}

\newtheorem{definition}{Definition}
\newtheorem{proposition}{Proposition}

\title{
Is a Linear Probe Evidence of a Linear Representation?
}

\author{
  Nadir Kutluozen$^{*}$ \\
  Cold Spring Harbor Laboratory \\
  \And
  Christian Intern\`{o}$^{*}$ \\
  Bielefeld University \\
  \And
  David Klindt \\
  Cold Spring Harbor Laboratory \\
}
\begin{document}
\maketitle
\vspace{-3em}
\begin{center}
 $^{*}$ \textit{Equal contribution}
\end{center}
% \vspace{0.2em}
{\renewcommand{\thefootnote}{}\footnotetext{\hspace{-1.8em}Correspondence: \texttt{nadirkutluozen.nk@gmail.com}, \texttt{christian.interno@uni-bielefeld.de}. \\ Code at \url{github.com/klindtlab/linear-probe-nullspace}}}
\etocdepthtag.toc{mtmain}
% \vspace{-1.2em}
% \begin{center}
% \small
% \faGlobe~\href{...}{\texttt{...}}\qquad
% \faGithub~\href{github.com/klindtlab/linear-probe-nullspace}{\texttt{code}}
% \end{center}
% ============================================================================
\begin{abstract}
The \textit{linear representation hypothesis} holds that pretrained vision models encode generative factors as linearly decodable directions in feature space, and comes with a standard methodology: train a linear probe, declare a factor \textit{represented} if accuracy is high. We show this test is too permissive. Asymptotically, any sufficiently rich feature map, including a random one, admits a linear readout, so high probe accuracy is consistent with two very different geometries: a true linear world model, where the latent-aligned subspace dominates the representation, or a kernel machine, where the latents lie linearly inside a much larger curved manifold. This distinction matters because the curved structure orthogonal to the probe is relevant for out-of-distribution generalization, robustness, and compositional transfer. We propose three asymmetric diagnostics that distinguish the two cases: reverse predictivity, spectral concentration, and out-of-distribution generalization. We validate them in a fully observed simulation where ground truth is known. Applied to six pretrained vision transformers on SVG-World, our deterministic generative benchmark with paired visual styles, and to two of them on CelebA, they show pretraining produces structure beyond a random kernel but well short of a linear world model. On dSprites, 3D Shapes and MPI3D, forward decodability and latent-aligned variance dissociate. Thus, forward decoding alone does not show that the probed factors align with the dominant geometry of the representation. In summary, our work cautions about the success of linear probes as opposed to truly linear and low dimensional representations.
\end{abstract}

\begin{figure}[ht]
\centering
\includegraphics[width=0.95\textwidth]{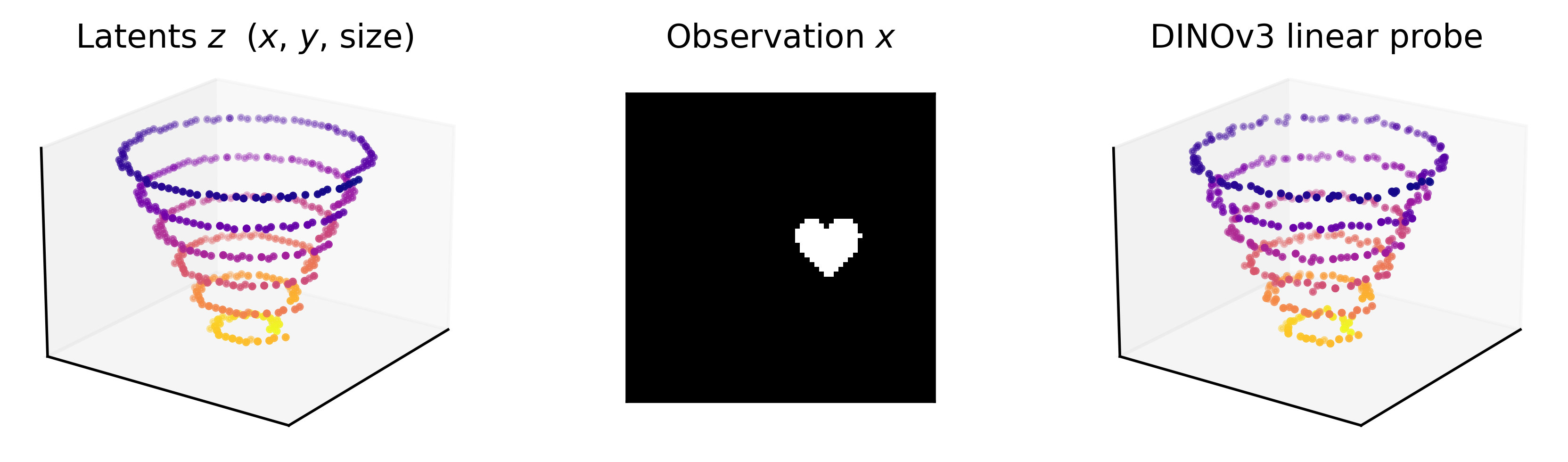}
\caption{\textbf{Forward Linear Decoding.} A 3D latent trajectory $z$ (left) generates images $x$ (middle) through a renderer \citep{matthey2017dsprites}. A linear probe on DINOv3-B/16 features recovers the trajectory in latent space (right), inverting the rendering pipeline with a linear map. We ask when recovery is evidence of a linear world model (Def.~\ref{def:lwm}) rather than
 a generic property of any sufficiently rich feature map.}
\label{fig:hook}
\end{figure}

% ============================================================================
\section{Introduction}
\label{sec:intro}

A recurring claim in representation learning is that pretrained models internally organize the world into linearly accessible factors of variation \citep{alain2017understanding}. This \emph{linear representation hypothesis} has been invoked to explain phenomena ranging from analogies in word embeddings \citep{mikolov2013linguistic, rumelhart1986distributed} to mechanistic interpretability of large language models \citep{park2023linear, elhage2022toymodels}, and it underwrites an increasingly popular methodology: train a linear probe and declare a latent variable ``represented'' when $R^2$ is high. The implicit inference is: \emph{if $z$ is linearly decodable from $f(x)$, then $f$ has learned $z$ as an axis of its representation.} We formalize the strongest version of this hypothesis, namely, that $z$ spans the dominant directions of $f$'s variance, as a \emph{linear world model} of $z$ (Def.~\ref{def:lwm}).

% ============================================================================
% Figure 1: hook
% ============================================================================

Testing this hypothesis is more nuanced than it appears. 
Rich nonlinear feature maps, including random ones, are universal function approximators, so with enough features almost any sufficiently smooth target can be linearly read out from them \citep{rahimi2007random}. The infinite-width limit at initialization is itself a Gaussian process \citep{neal1996bayesian, lee2018deep}, and the neural tangent kernel \citep{jacot2018ntk} extends this to wide networks during training.
A high forward $R^2$ is therefore consistent with two distinct hypotheses: (\textit{H1}) $f$ is a linear world model of $z$ (Def.~\ref{def:lwm}), or (\textit{H2}) the model is a high-dimensional feature map in which $z$ lies linearly but carries a small share of the variance. Empirically, probes are known to recover signal from models that lack the structure \citep{hewitt2019designing, belinkov2022probing, levinstein2025lie, internò2026observereffectworldmodels}; we extend this concern from the probe to the encoder itself. Probe success is also subject to a failure mode under \textit{superposition}, where linear representation is necessary but not sufficient for recovery \citep{pacela2026stopprobingstartcoding, klindt2025superposition}.

\textbf{Contributions.} We \textbf{(i)} formalize the notion of a \textit{linear world model} (Def.~\ref{def:lwm}) and propose three asymmetric diagnostics, reverse predictivity, spectral concentration, and out-of-distribution (OOD) generalization, each of which can fail for a kernel-machine map that satisfies forward decoding, plus a ceiling that makes reverse predictivity comparable across models (\S\ref{sec:theory}); \textbf{(ii)} validate them in a fully observed simulation where 2D latents are lifted by a frozen random Fourier map into random or trained MLP features, recovering the asymptotic random-features prediction for the random-init baseline and the Def.~\ref{def:lwm} regime for the trained, demonstrably nonlinear counterpart (\S\ref{sec:simulation}); \textbf{(iii)} introduce SVG-World, a deterministic testbed with a 32-dim latent layout shared between two rendering styles (AquaWorld, WestWorld), and apply the diagnostics to six pretrained vision transformers from three training paradigms (DINOv3 S/B/L, DINOv2, DeiT, CLIP; DINOv3-B/16 and CLIP-B/16 also on dSprites, 3D Shapes, MPI3D and CelebA), finding that DINOv3-B/16 sits partway between the simulation's two regimes with style-invariant alignment concentrated in the top $\sim$32--64 PCs (\S\ref{sec:svg-world}--\ref{sec:breadth}).

% ============================================================================
\section{Related Work}
\label{sec:related}

\textbf{Linear Representations and World Models.} From word-embedding analogies \citep{mikolov2013linguistic} and earlier distributed representations \citep{rumelhart1986distributed}, through vision probes \citep{alain2017understanding}, to recent mechanistic interpretability of LLMs \citep{park2023linear, elhage2022toymodels}, a line of work claims that high-level concepts are encoded as linear directions; sparse autoencoders \citep{cunningham2023sparse, bricken2023monosemantic} add that linearity plus sparsity recovers the ``atoms'' of the representation under superposition. \emph{World model} has been used with varying formality, from latent dynamics models for RL \citep{ha2018worldmodels, hafner2020dreamer} to modern video generators \citep{brooks2024sora}, and the hope that pretrained models instantiate one \citep{lecun2022path} is often operationalized by linear probing on factors believed to be relevant.

\textbf{Identifiability, Disentanglement, and High-Dimensional Representations.} Nonlinear ICA \citep{hyvarinen2019nonlinear, khemakhem2020ica} and linear identifiability up to affine transforms \citep{roeder2021linear} give formal conditions for recovering latents; classic work shows disentanglement is unidentifiable without inductive biases or supervision \citep{hyvarinen1999nonlinear, locatello2019challenging}, and \citet{higgins2021symmetry} propose a symmetry-based definition. Our reverse-predictivity diagnostic \citep{muzellec2025reverse} is closely related to the covariance-alignment arguments of this literature. \citet{stringer2019high} show that neural responses follow a power-law spectrum, which the strict ``variance-concentration'' reading of Def.~\ref{def:lwm} cannot satisfy (\S\ref{sec:discussion}).

\textbf{Random Features and Kernels.} \citet{rahimi2007random} show that random feature maps approximate RKHS kernels, so arbitrarily complex targets become linearly recoverable with enough dimensions; the neural tangent kernel \citep{jacot2018ntk} extends this to wide networks at initialization, and later work relates the kernel eigenspectrum to learnability and generalization \citep{bordelon2020spectrum, canatar2021spectral}. \citet{sprekeler2014} make the same point for nonlinear ICA: random polynomial features with slow feature analysis recover sources without learning the feature map. All predict an asymptotic equivalence of trained and random feature maps for forward linear decoding; characterizing the finite-$d$ regime is part of our contribution.

\textbf{Generalization-Based Diagnostics.} A representation that genuinely encodes a factor should support compositional generalization \citep{lake2018scan, keysers2020compositional} and extrapolation beyond the training support. \citet{schott2022visual} report that pretrained representations often fail to generalize compositionally even within one domain, so within-support forward decoding can mislead. The cross-condition generalization performance of \citet{bernardi2020geometry} operationalizes a related idea in neuroscience, i.e., transfer of linear separability; we adopt a related framing in \S\ref{sec:ood}, with two paired visual styles as the conditions.

% ============================================================================
\section{Theory: When Is Linear Decoding Evidence of a Linear World Model?}
\label{sec:theory}

% --- Figure: linear world model vs kernel machine, concept and simulation realisation ---
\begin{figure}[ht]
\centering
\includegraphics[width=0.99\linewidth]{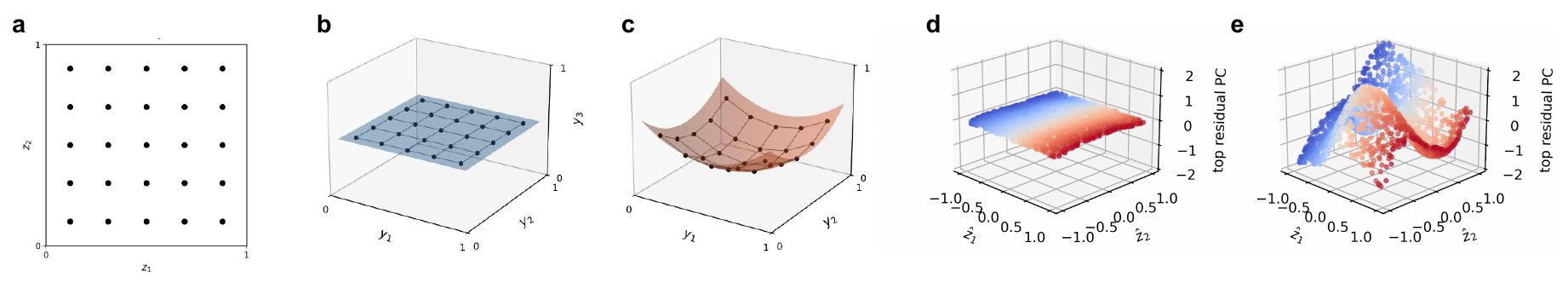}
\caption{\textbf{Linear World Model vs.\ Kernel Machine.} \textbf{(a)} 2D latent grid. \textbf{(b, c)} Cartoons: a linear world model lifts the grid to a flat plane (Def.~\ref{def:lwm}); a kernel-machine map might lift it to a curved surface. Forward $R^2$ is high in both; reverse $R^2$ separates them (Prop.~\ref{prop:reverse}). \textbf{(d, e)} Empirical realization at $h{=}1024$: horizontal axes are forward-decoded $\hat z_1, \hat z_2$; vertical is the top variance direction in the decoder's null space, residualized against $\hat z$. Trained is a plane, random a curved manifold.}
\label{fig:simulation-3d}
\end{figure}

We begin with a formal definition, recall the asymptotic argument that makes forward decoding a weak diagnostic in the limit, and close with three asymmetric (dimension independent) tests.

\subsection{Setup and Definition}
Let $\X$ be a space of inputs and $z : \X \to \R^k$ the latents (of the \textit{data generating process}). A frozen feature map $f : \X \to \R^d$ produces embeddings on which we run linear probes.

\begin{definition}[Linear world model]
\label{def:lwm}
We say that $f$ \emph{implements a linear world model of $z$ with residual level $\sigma^2 \ge 0$} if, with $\tilde z = \Cov(z)^{-1/2}(z - \E [z])$ the whitened latents, there exist $W \in \R^{d \times k}$ and $\mu \in \R^d$ such that
\begin{equation}
f(x) = W \tilde z(x) + \mu + \varepsilon(x),
\qquad \E\!\left[\varepsilon(x)\, \tilde z(x)^\top\right] = 0,
\qquad \E\!\left[\norm{\varepsilon(x)}^2\right] \le \sigma^2 \norm{W}_F^2.
\end{equation}
\end{definition}

That is, $f$ is, up to a bounded orthogonal residual, an affine map of $z$. The residual $\varepsilon$ is the nonlinear part of $f$, for example curvature in $z$ or variation due to other factors. Since $\Cov(\tilde z) = I$, $\norm{W}_F^2$ is the variance of $f$ carried by the latents, so $\sigma^2$ is the ratio of residual to latent-aligned variance (unit-free), and $\sigma^2 = 0$ means $f$ is exactly an affine image of $\tilde z$. As with linear probes, we call the map linear: $\mu$ drops out of every statistic below, all computed on centered data. Throughout, ``linear world model'' means only Def.~\ref{def:lwm}, variance-dominant linear alignment with a specified factor set, not a general account of how $f$ represents the world. This is strictly stronger than ``$z$ is linearly decodable from $f$'': it requires that \emph{most} of $f$'s variance is aligned with $z$, not merely that $z$ lies inside the span of $f$. Def.~\ref{def:lwm} is one end of an axis whose other end is a rich feature map that merely admits a linear readout of $z$ (Prop.~\ref{prop:rf}); the diagnostics locate $f$ on this axis for a specified factor set.

\subsection{Forward Decodability Is Asymptotically Free}

The following statement is a direct consequence of classical random-features theory \citep{rahimi2007random}; we state it informally and defer the technical version to App.~\ref{app:theory}.

\begin{proposition}[Forward decodability from rich random features]
\label{prop:rf}
Let $\kappa(\cdot,\cdot)$ be a bounded continuous kernel on $\X$ whose RKHS $\Hc_\kappa$ contains every coordinate $z_j$ of the ground-truth latents. Let $f_\mathrm{rand}(x) \in \R^d$ be a random feature map such that $f_\mathrm{rand}(x)^\top f_\mathrm{rand}(x') \to \kappa(x,x')$ as $d \to \infty$. Then for any $\epsilon > 0$ and $\delta \in (0,1)$ there exists $d_{\epsilon,\delta} \in \mathbb N$ such that for all $d \ge d_{\epsilon,\delta}$, with probability at least $1-\delta$ over the draw of $f_\mathrm{rand}$, there exists $W \in \R^{k \times d}$ with
\begin{equation}
\E_x\!\left[\norm{W f_\mathrm{rand}(x) - z(x)}^2\right] \le \epsilon.
\end{equation}
\end{proposition}

Prop.~\ref{prop:rf} has an asymptotic consequence: forward linear decodability is, in the limit, a property of having enough random features rather than of having learned anything about $z$. Whether the asymptotic regime dominates at the feature dimensions used in practice ($d$ on the order of $10^2$ to $10^3$) is empirical, and we test it directly in \S\ref{sec:simulation} and \S\ref{sec:dino-forward}.

\subsection{Three Asymmetric Diagnostics}
\label{sec:diagnostics}

We propose three tests, each of which exploits a different asymmetry between Def.~\ref{def:lwm} and the mere existence of a linear decoder.

\paragraph{(D1) Reverse predictivity \citep{muzellec2025reverse}.} Under Def.~\ref{def:lwm}, the latent-aligned subspace accounts for a large share of feature variance. We therefore measure
\begin{equation}
R^2_\mathrm{rev}
\;=\;
1 - \frac{\E_x \norm{f(x) - W_\mathrm{rev} z(x) - \mu}^2}{\E_x \norm{f(x) - \E f}^2},
\qquad
W_\mathrm{rev} = \argmin_{W,\mu} \E_x \norm{f(x) - W z(x) - \mu}^2.
\end{equation}

\begin{proposition}[Reverse-predictivity asymmetry]
\label{prop:reverse}
Suppose $f(x) = A z(x) + \mu + \varepsilon(x)$ with $\E[\varepsilon z^\top] = 0$. Then
\begin{equation}
R^2_\mathrm{rev}
\;=\;
\frac{\tr(A \Cov(z) A^\top)}{\tr(A \Cov(z) A^\top) + \tr \Cov(\varepsilon)}.
\end{equation}
In particular, if $\varepsilon$ is dominated by a high-dimensional $z$-uncorrelated nonlinear component (the kernel-machine regime), $R^2_\mathrm{rev}$ is small even when forward $R^2 \approx 1$.
\end{proposition}

\begin{proposition}[Ceiling]\label{prop:ceiling}
For any $f$, let $\lambda_1 \ge \dots \ge \lambda_d$ be the eigenvalues of $\Cov(f)$ and $r = \rank\Cov(z)$. Then
\begin{equation}
R^2_\mathrm{rev} \;\le\; C_r \;=\; \frac{\sum_{i \le r} \lambda_i}{\sum_{i \le d} \lambda_i},
\end{equation}
with equality iff the range of $W_\mathrm{rev}$ lies in a leading $r$-dimensional principal subspace of $f$ and the reverse-probe residual $f - W_\mathrm{rev} z$ has no variance in that range.
\end{proposition}
The proof is in App.~\ref{app:theory}. We use $r = \rank\Cov(z)$ rather than $k$ so that constant or redundant latent coordinates do not loosen the bound. The ratio $R^2_\mathrm{rev}/C_r$ makes intermediate values interpretable: it is the fraction of the attainable share that the factors take, comparable across encoders with different spectra. For a scalar factor, $C_1 = \lambda_1/\tr\Cov(f)$ bounds even a perfectly isolated factor. Since $C_r$ depends on the spectrum, we also report the participation ratio $\mathrm{PR} = (\sum_i \lambda_i)^2/\sum_i \lambda_i^2$.

\textbf{(D2) Spectral concentration.} Under Def.~\ref{def:lwm} with small $\sigma^2$, the latent-aligned subspace is spanned by the leading principal directions of $f$. D2 asks whether this is the case, independently of how much variance those directions carry. Project $f$ onto its top $m$ principal components and rescale each to unit variance (PCA whitening, fitted on training rows only), and let $R^2_\mathrm{rev}(m)$ be the reverse $R^2$ of the result. Because the whitened components have equal variance, $R^2_\mathrm{rev}(m)$ is the average, over the top $m$ principal components, of the fraction of each component's variance that is linear in $z$. We report the scalar $\mathrm{SC} = R^2_\mathrm{rev}(r)$, with $r = \rank\Cov(z)$ as in Prop.~\ref{prop:ceiling}: the extent to which the $r$ leading directions of $f$ are linear functions of the latents. A strict linear world model has $\mathrm{SC} \approx 1$. A kernel machine, whose leading directions are dominated by $z$-uncorrelated variation, has low SC even when forward $R^2$ is high. For $m > r$ the curve falls for any $f$, at least as $r/m$ once the extra components are residual, so the informative region is $m \approx r$; the full curve shows where in the spectrum the latent-aligned variance lives. Unlike subspace-similarity measures such as CKA \citep{kornblith2019similarity}, SC needs no choice of basis for the latent-aligned subspace, only its location in $f$'s spectrum.

\textbf{(D3) Out-of-distribution generalization.} Linear decodability that holds \emph{only within training support} is consistent with overfitting a rich nonlinear kernel on that support. A genuine linear world model should extrapolate \citep{lachapelle2023additive}. We test this by training linear probes on one visual-style condition and evaluating on another condition with similar latents; the encoder's positional directions should transfer if it has actually learned position as a factor rather than as a style-conditional pattern \citep{lake2018scan, schott2022visual, bernardi2020geometry}.

\textbf{Summary.} For a linear world model with small $\sigma^2$, all three diagnostics are close to their maximum: $R^2_\mathrm{rev}$ and SC are near $1$, and a probe trained in one condition transfers to another with little loss. For a rich kernel-machine map, forward $R^2$ can be high while all three are low. The diagnostics are therefore necessary conditions: low scores rule out a linear world model of $z$, high scores do not establish one. Like Def.~\ref{def:lwm}, they are linear, so a factor encoded faithfully along a curved manifold also scores low. The rest of the paper asks where pretrained vision models fall between these two cases, for a specified factor set.

% ============================================================================
\section{Controlled Validation in a Fully Observed Simulation}
\label{sec:simulation}

% --- Figure: simulation performance sweep ---
\begin{figure}[b]
\centering
\includegraphics[width=\linewidth]{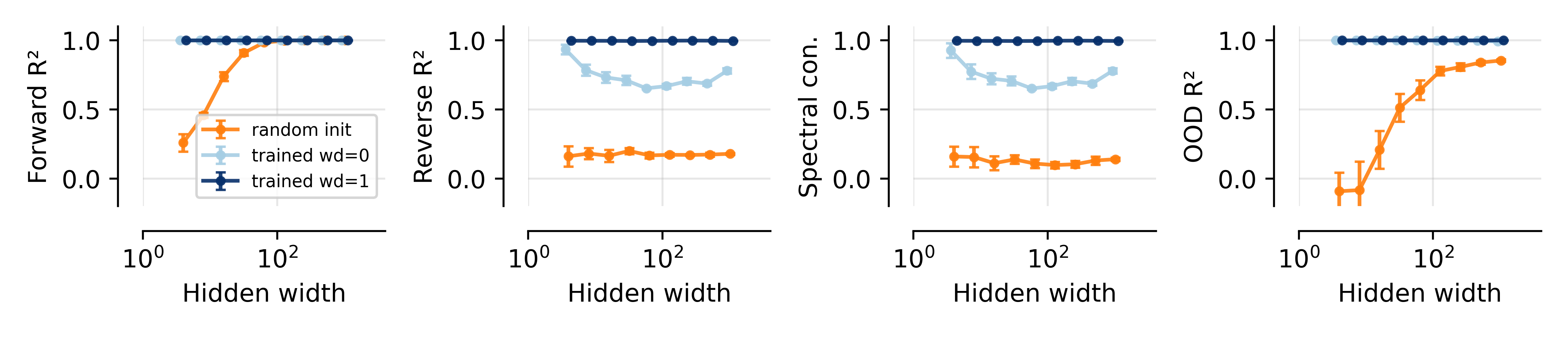}
\vspace{-10pt}
\caption{\textbf{Diagnostics on the simulation across hidden width $h$.} Random init (orange) reproduces the asymptotic random-features prediction: forward $R^2 \to 1$ as $h$ grows, while reverse $R^2$ and the top-2 share stay low and OOD $R^2$ catches up only at large widths. Trained with weight decay $\lambda = 1$ (dark blue), all four diagnostics sit flat at $\approx 1$ across widths. Without weight decay (light blue), reverse $R^2$ and the top-2 share dip at intermediate widths: the linear-world-model condition is sensitive to training pressure even when forward $R^2$ saturates. The third panel is the share of variance in the top two PCs that is linear in $z$, without whitening (App.~\ref{app:simulation}). Error bars over 3 seeds (initialization and batch order).}
\label{fig:simulation-perf}
\vspace{-10pt}
\end{figure}

Before turning to vision foundation models, we verify that the diagnostics of \S\ref{sec:diagnostics} behave as predicted in a setting where ground truth is known and the random-features prediction can be tested directly.

\textbf{Setup.} Latents $z \sim \mathrm{Unif}([-1, 1]^2)$ are first lifted by a frozen random Fourier map, $x_j = \cos(w_j^\top z + b_j)$ for $j = 1, \dots, 256$, with $w_j \sim \mathcal N(0, 3^2 I_2)$ and $b_j \sim \mathrm{Unif}[0, 2\pi]$ drawn once and never trained \citep{rahimi2007random}, then mapped to features $f(z) \in \R^h$ by a two-layer ReLU MLP, $f(x) = W_2\,\mathrm{ReLU}(W_1 x + b_1) + b_2$ \citep[c.f., ][]{sprekeler2014}. We compare two regimes at matched hidden width $h$: (i) \emph{random init}, fresh MLP weights, no training; (ii) \emph{trained}, the same architecture trained end-to-end with a linear readout $Uf$ to recover $z$ using AdamW (learning rate $2\cdot 10^{-3}$, 2000 steps, batch 256), weight decay $\lambda \in \{0, 1\}$.We hypothesize that weight decay encourages linear world models: reading $z$ off a low-variance direction of $f$ needs a large readout row, so penalizing $\|U\|$ pushes the $z$-aligned subspace toward the leading directions. The network never sees $z$, and it is nonlinear: at $h = 1024$ (5 seeds, 2000 test points) it visits 1988 ReLU activation regions on average and its Jacobian varies across inputs, where an affine network has one region and a constant Jacobian (App.~\ref{app:simulation}).

\textbf{Result 1: Forward Decoding Alone Does Not Separate the Two.} Fig.~\ref{fig:simulation-perf} (leftmost panel) shows the asymptotic prediction of Prop.~\ref{prop:rf} directly: random-init forward $R^2$ rises smoothly with width, reaching $\approx 1$ by $h \approx 100$. By that width, the random feature map is forward-decodable as well as the trained model. Forward $R^2$ alone cannot distinguish the two beyond very small widths.

% --- Figure: simulation spectral analysis ---
\begin{figure}[t!]
\centering
\includegraphics[width=\linewidth]{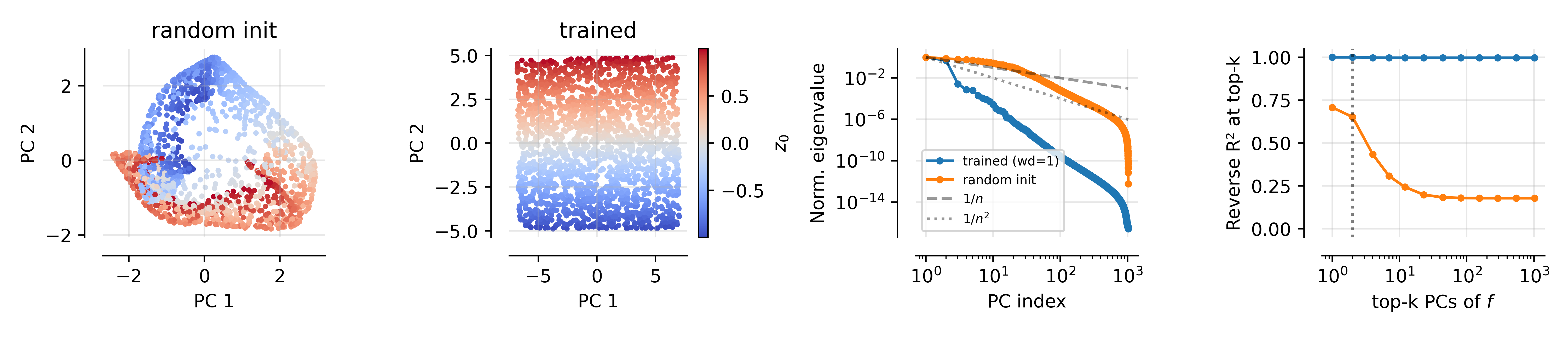}
\vspace{-20pt}
\caption{\textbf{Spectral diagnostics on the simulation} ($h = 1024$; random init vs.\ trained with $\lambda = 1$). (a, b) Test points on the top-2 PCs of $f$, colored by $z_0$. Random folds the latent square into a non-axis-aligned shape (top-2 subspace not $z$-aligned); trained gives an axis-aligned image, $z_0$ along PC 2 and $z_1$ along PC 1. (c) Normalized eigenspectrum of $f$ (log-log): trained drops by several orders of magnitude after PC 2, random decays smoothly over hundreds of PCs. (d) Reverse $R^2$ on the top $m$ PCs of $f$, without whitening. Trained stays at 1: its top two PCs are linear in $z$ and carry almost all the variance, so later PCs barely change the ratio. Random falls from $\approx 0.7$ at $m = 1$ to a plateau at its full $R^2_\mathrm{rev} \approx 0.2$ as PCs not linear in $z$ are added.}
\label{fig:simulation-spectrum}
\vspace{-10pt}
\end{figure}

\textbf{Result 2: The Asymmetric Diagnostics Open a Robust Gap.} Reverse $R^2$ and the top-2 share separate trained from random at every width, and OOD at small and moderate widths (Fig.~\ref{fig:simulation-perf}, right three panels). The trained-saturated curve (weight decay $\lambda = 1$) sits at $1$ on all three; random sits well below it on reverse $R^2$ and the top-2 share, and approaches it on OOD only at large widths. Prop.~\ref{prop:reverse} predicts random's reverse $R^2$ should approach $0$ as the $z$-orthogonal residual dimension grows; the empirical curve plateaus near $0.2$ rather than decaying, an artifact of finite-width ReLU MLPs not realizing the strict $z$-orthogonality the proposition assumes. The middle curve ($\lambda = 0$, supervised but unregularized) is substantively worse than the saturated case on reverse $R^2$ and the top-2 share, sagging to $\sim 0.7$ at intermediate widths, and is indistinguishable from the saturated case on forward and OOD.  Consistent with this hypothesis, at $h = 1024$ (5 seeds), reverse $R^2$ is $0.9935$ with $\lambda = 1$ vs.\ $0.7841$ with $\lambda = 0$, at identical forward $R^2$ of $1.000$. \emph{We conclude that satisfying Def.~\ref{def:lwm} is a tunable property of training pressure, not just a binary trained-vs-not property.}

\textbf{Result 3: A Linear World Model Is Visible in Feature Space.} Fig.~\ref{fig:simulation-3d}(d, e) plots random and weight-decay ($\lambda = 1$) models at $h = 1024$ against the forward-decoded latents and the largest residual direction. For the trained model the points lie on a flat plane, an affine image of the latent square, as Def.~\ref{def:lwm} predicts. For the random model they lie on a saddle: $z$ is linearly decodable, but the feature map also bends the latent square along directions the probe ignores. This curvature is the residual variance that lowers $R^2_\mathrm{rev}$ in Prop.~\ref{prop:reverse}.

\textbf{Result 4: The Spectral Picture.} Fig.~\ref{fig:simulation-spectrum} compares the spectra of the random model and the model trained with $\lambda = 1$. The trained covariance is nearly rank two: the top two components carry $99.57\%$ of the variance (shares $0.5657$ and $0.4300$, against $0.1384$ and $0.1151$ at random init), the eigenvalues drop by several orders of magnitude after the second (panel c), and the projection onto the top two PCs is an axis-aligned image of the latent square (panel b). The random model's spectrum decays smoothly over hundreds of components, and its top-two projection folds the latent square onto itself (panel a). Panel (d) shows the reverse $R^2$ after projecting $f$ onto its top $m$ PCs, without whitening. For the trained model it stays at $1$ for every $m$, because the leading PCs are linear in $z$ and the later ones carry almost no variance. For the random model it falls from $\approx 0.7$ at $m = 1$ and levels off at $\approx 0.2$, its full $R^2_\mathrm{rev}$: each added PC brings variance that is not linear in $z$, which is the dilution described by Prop.~\ref{prop:reverse}.

\textbf{Takeaway for the Empirical Sections.} The simulation verifies that the diagnostics behave as Def.~\ref{def:lwm} and Prop.~\ref{prop:reverse} predict, and that forward $R^2$ alone is insufficient to discriminate the two regimes once $d$ is moderately large. The remaining empirical question is where DINOv3 sits on this scale.

% ----- Figure: SVG-World latents -----
\begin{figure}[t]
\centering
\includegraphics[width=0.9\linewidth]{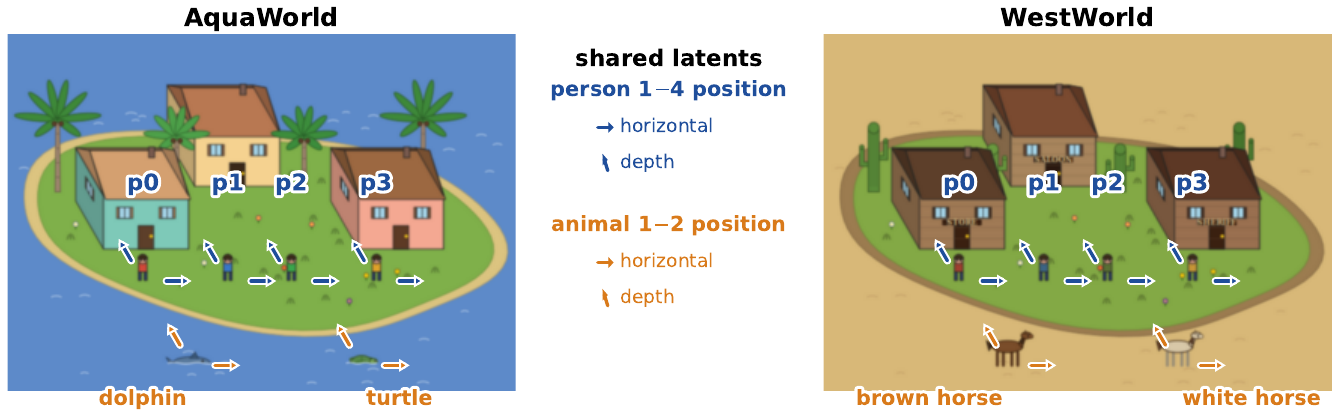}
\caption{\textbf{SVG-World, a deterministic two-style benchmark with a shared latent layout.} Both panels show the canonical scene at $z = 0$; arrows mark each object's two position latents ($+x$ horizontal, $+z$ depth, foreshortened by the oblique projection). AquaWorld  and WestWorld  share the same 32-dim layout (4 people with $(x, z)$ position, facing direction and skin hue; 2 animals with $(x, z)$ position) but differ in rendering, which enables the cross-domain transfer test of \S\ref{sec:ood}.}
\label{fig:svg-world}
\end{figure}

% ============================================================================
\section{SVG-World: A Controllable Two-Style Benchmark}
\label{sec:svg-world}

To test for the presence of linear representations in real models, we need a \textit{data generating process} (DGP), i.e., a simulation with fully specified latent variables. More precisely, we want a data source with (a) a small, enumerable set of continuous latents, (b) deterministic rendering, (c) enough visual richness that frozen vision models produce nontrivial features, (d) multi-object scenes for binding tests, and (e) a paired second visual style with the same latent layout, so that linear probes can be tested for style-invariance. Existing disentanglement benchmarks \citep{matthey2017dsprites, burgess20183dshapes} satisfy (a)--(c) but not (d) or (e). SVG-World is built to cover all five.

\textbf{Pipeline.} Each scene is rendered from a 32-dim latent vector $z \in [0, 1]^{32}$, of which 20 dimensions are active: 4 people each contribute a position $(x, z)$, a discrete facing direction, and a skin-hue value, and 2 animals each contribute a position $(x, z)$. The active latent layout is identical across both worlds (Fig.~\ref{fig:svg-world}); only the rendering differs. The rasteriser is \texttt{cairosvg} at $224 \times 224$ to match DINOv3's patch size. We release the AquaWorld and WestWorld splits ($N = 10{,}000$ each; \url{github.com/klindtlab/linear-probe-nullspace}); a dataset card with details is in App.~\ref{app:svg-world}.

\textbf{Why Two Paired Worlds.} The shared layout lets us run the same linear probe across both rendering styles and ask not just whether a latent is decodable, but whether the directions used to decode it are stable under a substantial visual-style change. We exploit this in \S\ref{sec:ood}.

\textbf{Empirical Scope.} For all empirical sections that follow we report metrics on the 12 position latents (8 person-position dims + 4 animal-position dims). A precursor benchmark with 15 single-object latents in continuous color space (SVG-Faces) is included as a consistency check in App.~\ref{app:svg-faces}.

% ============================================================================
\subsection{DINOv3 Forward Decoding}
\label{sec:dino-forward}

% ----- Figure: random-init comparison -----
\begin{figure}[t]
\centering
\includegraphics[width=0.7\linewidth]{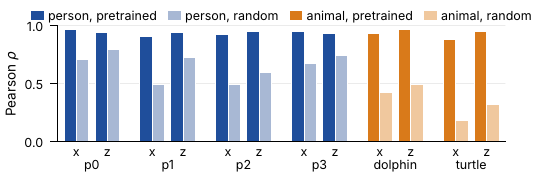}
\caption{\textbf{Forward Decodability Separates Pretrained from Random-Init at $d{=}768$.} Per-latent $\rho$ on CLS features; per-latent $R^2$ in Table~\ref{tab:dino-forward} (App.~\ref{app:experiments}). Pretrained averages $\rho = 0.94$ ($R^2 = 0.88$); random averages $0.55$ ($0.33$) ($\Delta\rho = 0.38$, rising to $0.58$ on animals).}
\label{fig:forward-random}
\vspace{-5pt}
\end{figure}

\textbf{Setup.} We extract frozen features from DINOv3-B/16 \citep{simeoni2025dinov3, caron2021dino} on each of the AquaWorld and WestWorld splits ($N = 10{,}000$ each, $50/50$ train/test), in two feature modes: the CLS token ($d = 768$) and the mean of patch tokens ($d = 768$). For each mode we fit ridge regression from $f(x)$ to each latent and report test $\rho$ and $R^2$ per latent, averaged over the two worlds. As a baseline we use \emph{random-init} DINOv3: the same architecture with freshly initialized weights, no training, and the same probing protocol. Implementation details (checkpoint, ridge penalty, seeds) are in App.~\ref{app:experiments}.

\textbf{Result.} Pretrained DINOv3 recovers all 12 position latents with mean $\rho \approx 0.94$ ($R^2 = 0.88$) (CLS) and $0.91$ ($0.82$) (mean-pooled patches), the kind of evidence routinely cited for the linear representation hypothesis (Table~\ref{tab:dino-forward}). Mean-pooling helps small foreground objects (animal $R^2$ $0.87 \to 0.91$) and hurts large ones ($R^2$ $0.89 \to 0.77$), as expected when patch pooling preserves spatial information that CLS aggregates away. Random-init partially decodes person positions ($R^2 = 0.41$) but loses animals almost completely ($R^2 = 0.17$): the pretraining contribution is concentrated on small foreground objects where feature richness alone does not suffice. These values use a fixed ridge penalty, which penalizes the lower-variance random features more strongly; with a cross-validated penalty random init reaches $R^2 = 0.66$ (Table~\ref{tab:encoders}), but the gap on animals remains.

\textbf{Pretraining, Not Feature Richness, Drives Forward Decoding.} Figure~\ref{fig:forward-random} visualizes the per-latent CLS comparison from Table~\ref{tab:dino-forward}. The asymptotic random-features prediction does not dominate at $d = 768$: pretrained recovers most position latents at $\rho > 0.9$ (all at $R^2 \ge 0.76$), while random-init is consistently below, and goes below $0.4$ ($R^2 \le 0.31$) on the animal positions. Per \S\ref{sec:simulation}, however, this forward gap does not by itself establish that DINOv3 satisfies Def.~\ref{def:lwm}. The asymmetric diagnostics in \S\S\ref{sec:reverse} and~\ref{sec:ood} test the structural claim directly.

% ============================================================================
\subsection{Reverse Predictivity and Spectral Concentration}
\label{sec:reverse}

\begin{figure}[ht]
\centering
\begin{minipage}{0.5\linewidth}
\includegraphics[width=\linewidth]{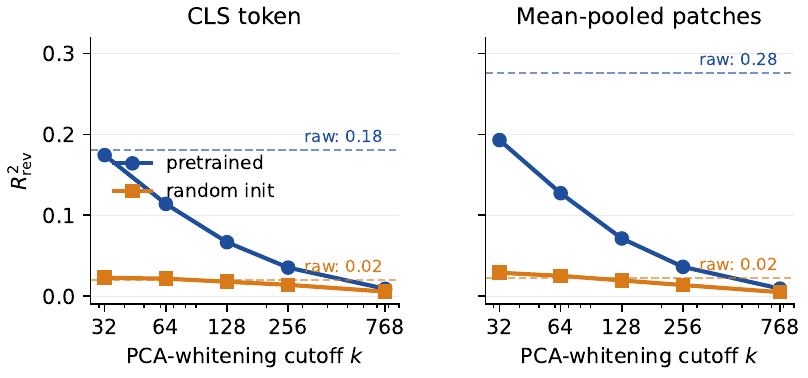}
\end{minipage}\hfill
\begin{minipage}{0.40\linewidth}
\caption{\textbf{Reverse Predictivity (D1) and Spectral Concentration (D2) on SVG-World.} $R^2_\mathrm{rev}$ vs.\ PCA-whitening cutoff $k$. Solid: top-$k$ PCs whitened to unit variance. Dashed: raw features. Pretrained's latent-aligned variance concentrates in the top $\sim$32--64 PCs; random's is diffuse. The curve falls with $k$ for any representation (App.~\ref{app:reverse-ablations}); what separates the two is its value at small $k$.}
\label{fig:reverse}
\end{minipage}
\vspace{-10pt}
\end{figure}

We instantiate (D1) reverse predictivity as defined in Prop.~\ref{prop:reverse} on DINOv3-B/16 features. On raw CLS features, pretrained DINOv3 reaches $R^2_\mathrm{rev} = 0.18$, vs.\ $0.02$ for random init; on mean-pooled patches, $0.28$ vs.\ $0.02$ (Fig.~\ref{fig:reverse}, dashed lines). Over five random initializations, the CLS gap is $0.187$ [$0.182$, $0.191$] in AquaWorld and $0.143$ [$0.139$, $0.148$] in WestWorld (95\% paired scene-bootstrap intervals). The pretrained-vs-random ratio is approximately $10\times$, much larger than the forward ratio ($\sim 1.4\times$ with a cross-validated penalty, App.~\ref{app:experiments}). \emph{The forward gap therefore understates how much more latent-aligned variance pretrained has}, while the reverse gap highlights it. Against the ceiling of Prop.~\ref{prop:ceiling} the separation is sevenfold: on CLS features in AquaWorld ($R^2_\mathrm{rev} = 0.212$, $C_{12} = 0.609$), pretrained reaches $0.35$ of the attainable share and random init $0.05$. Standardizing features per dimension before fitting the reverse probe leaves the numbers essentially unchanged (App.~\ref{app:reverse-ablations}), so the gap is not an artifact of variance scaling along individual feature dimensions.

\textbf{Spectral Concentration (D2).} The (D2) diagnostic asks where in the spectrum of $f$ the latent-aligned variance lives. We compute $R^2_\mathrm{rev}(k)$, the reverse $R^2$ after PCA-whitening features to their top $k$ principal components (every PC rescaled to unit variance), for $k \in \{32, 64, 128, 256, 768\}$. Figure~\ref{fig:reverse} shows the result. Pretrained's $R^2_\mathrm{rev}$ is largest at $k \approx 32$: $0.17$ on CLS and $0.19$ on patches. It decays monotonically as $k$ grows, falling below $0.04$ at $k = 256$ and to $0.01$ at full whitening ($k = 768$). Random init's curve is approximately flat at $0.02$ across all $k$.Pretrained DINOv3-B/16 needs its top 17 principal components to collect 90\% of its latent-predictable variance, against 55--59 for its random initializations (App.~D.1)

\textbf{Interpretation.} DINOv3's self-distillation objective applies no explicit pressure to make $f$ an affine image of any given factor set $z$; the encouragement of linear representations through weight decay in the simulation (\S\ref{sec:simulation}, Result 2) has no obvious analog in the SSL loss. Thus, we do not expect DINOv3 to reach the strict Def.~\ref{def:lwm} regime ($R^2_\mathrm{rev} \to 1$). What we do see is that pretraining produces clear spectral concentration of the moderate latent-aligned variance that exists, consistent with the spectral picture of the trained simulation in Fig.~\ref{fig:simulation-spectrum} but at a much lower absolute level. DINOv3 is therefore best read as occupying the middle of the kernel-machine to linear-world-model axis: well above a random-feature null on direction-alignment and spectral structure, but well below the Def.~\ref{def:lwm} regime.

% ============================================================================
\subsection{Out-of-Distribution Generalization Across Visual Styles}
\label{sec:ood}
\begin{figure}[ht]
\centering
\includegraphics[width=0.9\linewidth]{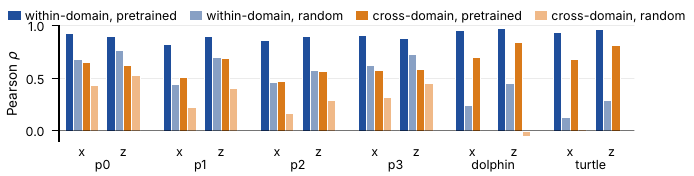}
\caption{\textbf{Asymmetric Diagnostic D3: Cross-Domain Transfer.} Per-latent test $\rho$ on DINOv3-B/16 mean-pooled patch features; cross-domain values average both transfer directions. The pretrained-vs-random gap survives the style change ($\Delta\rho = 0.41$ within, $0.42$ across domains). Animal positions transfer best (pretrained $\rho = 0.68$--$0.84$ cross-domain; random near zero).}
\label{fig:ood}
\end{figure}

We test (D3) by training a ridge probe on one world's features and evaluating it on the other's, with features mean-centered per world so what is tested is the transfer of \emph{directions}, not feature means. Figure~\ref{fig:ood} reports per-latent $\rho$ on DINOv3-B/16 mean-pooled patch features for both pretrained and random-init weights, in both within- and cross-domain conditions, averaged across both transfer directions (AquaWorld$\to$WestWorld and the reverse).

\textbf{Within-Domain vs.\ Cross-Domain.} Pretrained averages $\rho = 0.91$ within-domain vs.\ $0.51$ for random ($\Delta\rho = 0.41$). Cross-domain, the absolute values both fall (pretrained $0.65$, random $0.23$), but the gap between them does not shrink: $\Delta\rho = 0.42$ (with a cross-validated penalty the within-domain gap is $0.22$, so the gap grows under the style change). Pretraining's contribution is therefore style-invariant: the encoder's positional directions transfer across the AquaWorld $\leftrightarrow$ WestWorld stylistic gap as cleanly as they exist within a single world. Animal-position directions transfer particularly cleanly (cross-domain pretrained $\rho \approx 0.76$), while random-init falls to zero or negative on the same dims, despite both representations being trained or initialized identically across the two visual styles. We report $\rho$ because feature scales differ between the two styles, so uncalibrated transfer $R^2$ is not meaningful (it falls to $-58.8$). Rescaling the probe output on half of the target style gives transfer $R^2$ of $0.16$ and $0.31$ in the two directions for pretrained, against $0.01$ and $0.02$ for random init.

\textbf{Reading Alongside Reverse Predictivity.} (D1), (D2) and (D3) measure complementary properties of the encoder. Reverse $R^2$ asks whether $z$ accounts for a large share of $f$'s \emph{variance}; spectral concentration asks where in $f$'s spectrum that share lives; cross-domain transfer asks whether the directions $f$ uses to encode $z$ are style-conditional or style-shared. DINOv3 sits between a random-feature null on all three and the linear-world-model regime occupied by the trained simulation in \S\ref{sec:simulation}. Style-invariance is, however, the strongest signal: the size of the pretrained-vs-random gap survives a substantial change of rendering essentially undiminished, which is hard to reconcile with the kernel-machine hypothesis (H2, \S\ref{sec:intro}) but very natural under the latent-alignment hypothesis (H1). The next section tests whether these findings hold beyond DINOv3-B/16 and SVG-World.

\subsection{Beyond SVG-World: Other Encoders, Benchmarks and Real Photographs}
\label{sec:breadth}

\textbf{Setup.} We repeat the analysis with other encoders and datasets. On SVG-World we compare six vision transformers from three training paradigms: self-distillation (DINOv3 S/16, B/16, L/16 \citep{simeoni2025dinov3}; DINOv2-B/14 \citep{oquab2024dinov2}), supervised classification (DeiT-B/16 \citep{touvron2021deit}) and image--text contrastive learning (CLIP-B/16 \citep{radford2021clip}), each against a random initialization of the same architecture. DINOv3-B/16 and CLIP-B/16 are also evaluated on three standard disentanglement benchmarks, dSprites \citep{matthey2017dsprites}, 3D Shapes \citep{burgess20183dshapes} and MPI3D-realistic \citep{gondal2019mpi3d}, and on CelebA photographs \citep{liu2015celeba} (App.~\ref{app:experiments}). On these datasets we report forward $R^2$, reverse predictivity (D1) and spectral concentration (D2). The cross-style test (D3) requires two renderings of the same latent layout, which only SVG-World provides; on the benchmarks we instead hold out half of the joint combinations of two factors.

\begin{wrapfigure}{r}{0.33\linewidth}
\vspace{-15pt}
\centering
\includegraphics[width=\linewidth]{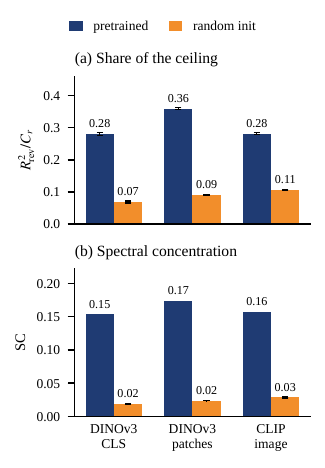}
\caption{\textbf{CelebA.} (a) Share of the ceiling, $R^2_\mathrm{rev}/C_{40}$, and (b) spectral concentration; pretrained vs.\ random init. Error bars: 95\% bootstrap for pretrained in (a), s.d.\ for random (Table~\ref{tab:celeba}).}
\label{fig:celeba}
\vspace{-15pt}
\end{wrapfigure}

\textbf{Encoders on SVG-World.}\label{sec:encoders} All six transformers have higher $R^2_\mathrm{rev}$ than their random initializations, by $0.145$ to $0.235$ (95\% intervals above zero). A ResNet-50 \citep{he2016resnet} shows a gap of $0.105$ (App.~\ref{app:experiments}). Within DINOv3, larger models decode better but carry less latent-aligned variance: from S/16 to L/16, forward $R^2$ rises from $0.855$ to $0.914$ while $R^2_\mathrm{rev}$ falls from $0.257$ to $0.169$.

\textbf{Controlled Benchmarks.}\label{sec:benchmarks} On dSprites and MPI3D the pretrained encoder decodes the factors better than its random initialization (forward $R^2$ $0.42$ vs.\ $0.16$ and $0.85$ vs.\ $0.61$) but has a lower $R^2_\mathrm{rev}$; on 3D Shapes forward $R^2$ is tied. The random initializations on these datasets are nearly low-dimensional (participation ratio at most $3.8$, against $8.7$ to $16.6$ pretrained), so most of their variance lies in the top $r$ components and their ceiling is close to $1$. % TODO: ceiling-normalized comparison for dSprites, 3D Shapes, MPI3D.
The comparisons also depend on how the factors are encoded: four of the six encoder--benchmark pairs keep the sign of the gap under all three encodings in App.~\ref{app:experiments}, two do not. % TODO: define the three encodings in App. D.

\textbf{Real Photographs.} On CelebA, pretrained models reach $0.28$ to $0.36$ of their ceiling against $0.07$ to $0.11$ at random initialization (Fig.~\ref{fig:celeba}, Table~\ref{tab:celeba}), close to the SVG-World values ($0.35$ vs.\ $0.05$, \S\ref{sec:reverse}). Spectral concentration separates them as clearly ($0.15$ to $0.17$ vs.\ $0.02$ to $0.03$). CLIP's raw $R^2_\mathrm{rev}$ exceeds that of DINOv3 CLS ($0.192$ vs.\ $0.170$), but both reach $0.28$ of the ceiling.

In summary, on SVG-World and CelebA pretraining raises the latent-aligned share well above random initialization, yet the best encoders reach only about a third of their ceiling, far from the linear-world-model regime of \S\ref{sec:simulation}; on the controlled benchmarks the comparison depends on encoder, dataset and factor encoding.

\section{Discussion}
\label{sec:discussion}

\textbf{Two Pictures Behind One Probe.} A successful linear probe is compatible with two quite different representations. In the first, the representation is essentially a low-dimensional subspace: its main directions of variance are affine in the latents, with little left over. In the second, the latents are one linear readout among many from a rich nonlinear expansion of the input (a kernel machine or random-feature model), and most of the variance is unrelated to them. Random-features theory makes the second picture forward-decodable for almost any smooth target once it is wide enough, so a probe alone cannot separate the two; (D1) reverse predictivity \citep{muzellec2025reverse}, (D2) spectral concentration and (D3) transfer across conditions can, because they ask about the representation, not the readout.

\textbf{Pretrained Encoders Fall in Between.} Across the encoders and datasets tested, wherever the comparison is unambiguous, pretrained representations fit neither picture. They carry far more latent-aligned variance than their random initializations, in the leading principal directions, and on SVG-World their position-decoding directions stay partially stable across visual styles. Yet the probed factors take at most about half of what any linear image of them could reach ($0.53$ at most), and within one model family larger encoders decode the factors better while devoting a smaller share of their variance to them. Pretraining thus adds latent-aligned structure to a rich feature map rather than replacing it with a low-dimensional one.

\textbf{Should We Expect Linear World Models?} Probably not, and the simulation suggests why: a network trained to report the latents became a linear world model only when its readout was penalized, which makes low-variance encodings of the latents costly; without the penalty it decoded as well but kept substantial variance off the latent subspace. Self-supervised objectives exert no comparable pressure towards an externally chosen factor set, and a general-purpose encoder serving many downstream tasks has good reason to represent more than any one; power-law spectra in visual cortex \citep{stringer2019high} point the same way for biological representations. The strict notion of Def.~\ref{def:lwm} is thus best read as one end of a scale, not a property to expect of pretrained models; the useful empirical question is how much of the representation a given factor set occupies, and where. Objectives explicitly favoring compressed or bottlenecked codes \citep{higgins2017beta, oord2018cpc} are the natural place to look for encoders closer to that end.

\textbf{What This Means for Probe-Based Claims.} Much of interpretability, and many claims that models contain world models, rest on the first picture: a probed direction is one of the axes along which the model organizes its inputs. Closer to the second picture, a probe can be accurate while the structure it ignores governs behavior out of distribution, under composition, or when the probed direction is used for an intervention. Like \citet{muzellec2025reverse}, we therefore suggest reporting, beside probe accuracy, reverse predictivity against its ceiling, where the latent-aligned variance sits in the spectrum, and transfer across conditions. The converse failure also occurs: under superposition, a representation linear in its factors can still defeat a linear probe \citep{pacela2026stopprobingstartcoding, klindt2025superposition}. Probe accuracy is thus neither necessary nor sufficient evidence for a linear representation, and the two failure modes call for different diagnostics.

\textbf{Scope and Limitations.} The diagnostics test whether a specified factor set carries the dominant directions of variance of a representation (Def.~\ref{def:lwm}). Low scores rule out a linear world model of those factors, not other structure, and our conclusions hold for the encoders and factor sets tested. Reverse predictivity depends on both: intermediate values must be read against the ceiling of Prop.~\ref{prop:ceiling}, and for two of six encoder--benchmark pairs the sign of the comparison depends on the factor encoding (\S\ref{sec:benchmarks}). Being linear and variance-based, reverse predictivity and spectral concentration underweight factors on sparse, low-variance or superposed directions, which transfer tests and a dictionary basis such as a sparse autoencoder detect better, and score a factor encoded faithfully along a curved manifold as low, which kernel-PCA or intrinsic-dimension controls would separate. CelebA extends the results to annotated photographs; open scenes without a specified factor set remain untested. Probes in \S\ref{sec:svg-world} and on CelebA use a fixed ridge penalty (cross-validated elsewhere), and the simulation uses a supervised loss rather than the self-supervised losses behind the encoders studied.

\textbf{Future Work.} Measuring how fast the forward gap between pretrained and random encoders closes as the feature dimension grows (Prop.~\ref{prop:rf}), for larger encoders and concatenated patch tokens, would separate feature richness from learned structure. The diagnostics apply unchanged to language models, including prompt-conditioned representations, and to published probes such as board-state \citep{li2023emergent, nanda2023emergent} or space and time \citep{gurnee2023language} directions, once their factor sets are specified. For steering and interventions \citep{li2023inferencetime, turner2023steering}, we expect directions from representations that fail the diagnostics to work in distribution but degrade under composition and shift, which is directly testable.

\begin{ack}
We would like to thank Habon Issa, Hazel Nam, Sunny Liu, Junze Zhou, Sonia Joseph and Liv Gorton for insightful discussions and feedback. 
Christian Internò acknowledges funding from the Honda Research Institute Europe.
David Klindt acknowledges the CSHL GPU cluster with assistance from the US National Institutes of Health Grant S10OD028632-01.
This work was supported by a Goodfire Research Grant and conducted using Silico, Goodfire’s interpretability research agent.
\end{ack}

% ============================================================================
\bibliographystyle{unsrtnat}
\bibliography{references}

% ============================================================================
\newpage
\appendix
\etocdepthtag.toc{mtappendix}
\etocsettagdepth{mtmain}{none}
\etocsettagdepth{mtappendix}{subsection}
\etocsettocstyle{\section*{Appendix Contents}}{}
\tableofcontents
\newpage
% ============================================================================
\section{Full Theoretical Statements and Proofs}
\label{app:theory}

This appendix gives the formal version of Prop.~\ref{prop:rf} and the proofs of Props.~\ref{prop:rf}, \ref{prop:reverse} and~\ref{prop:ceiling}.

\subsection{Proposition~\ref{prop:rf} (Forward Decodability from Rich Random Features)}

\paragraph{Formal statement.} Let $\X$ be a compact metric space with data distribution $p_\X$, and let $\kappa : \X \times \X \to \R$ be a continuous, bounded, positive-definite kernel with reproducing kernel Hilbert space $\Hc_\kappa$. Suppose each latent coordinate satisfies $z_j \in \Hc_\kappa$ with $\norm{z_j}_{\Hc_\kappa} \le R$. Let $\mathbb P_\omega$ be a distribution over parameters $\omega$ and $\psi_\omega : \X \to \R$ feature functions such that
\begin{equation}
\label{eq:rf-kernel}
\E_\omega\!\left[\psi_\omega(x)\,\psi_\omega(x')\right] \;=\; \kappa(x,x')
\qquad \text{for all } x,x' \in \X,
\end{equation}
with $\sup_x |\psi_\omega(x)| \le K$ almost surely. Draw $\omega_1,\dots,\omega_d$ i.i.d.\ from $\mathbb P_\omega$ and set $f_\mathrm{rand}(x) = d^{-1/2}\,(\psi_{\omega_1}(x),\dots,\psi_{\omega_d}(x))^\top \in \R^d$. Then for any $\epsilon > 0$ and $\delta \in (0,1)$ and all
\begin{equation}
d \;\ge\; d_{\epsilon,\delta} \;=\; \frac{k\,K^2 R^2}{\epsilon\,\delta},
\end{equation}
with probability at least $1-\delta$ over $\omega_1,\dots,\omega_d$ there exists $W \in \R^{k \times d}$ with
\begin{equation}
\E_{x \sim p_\X}\!\left[\norm{W f_\mathrm{rand}(x) - z(x)}^2\right] \;\le\; \epsilon.
\end{equation}

\paragraph{Proof.} Fix $j \in \{1,\dots,k\}$. Because $\kappa$ has the integral representation~\eqref{eq:rf-kernel}, every $g \in \Hc_\kappa$ can be written as $g(x) = \E_\omega[\beta(\omega)\,\psi_\omega(x)]$ for some $\beta \in L^2(\mathbb P_\omega)$, and $\norm{g}_{\Hc_\kappa}$ is the smallest $\norm{\beta}_{L^2(\mathbb P_\omega)}$ that achieves this \citep{rahimi2008weighted, bach2017equivalence}. Choose such a $\beta_j$ for $z_j$, so that
\begin{equation}
z_j(x) \;=\; \E_\omega\!\left[\beta_j(\omega)\,\psi_\omega(x)\right],
\qquad
\E_\omega\!\left[\beta_j(\omega)^2\right] \;\le\; R^2.
\end{equation}
Let $w_j = d^{-1/2}\,(\beta_j(\omega_1),\dots,\beta_j(\omega_d))^\top$, so that
\begin{equation}
\hat z_j(x) \;:=\; w_j^\top f_\mathrm{rand}(x) \;=\; \frac{1}{d}\sum_{i=1}^d \beta_j(\omega_i)\,\psi_{\omega_i}(x).
\end{equation}
For fixed $x$, $\hat z_j(x) - z_j(x)$ is an average of $d$ i.i.d.\ zero-mean terms, so
\begin{equation}
\E_{\omega_{1:d}}\!\left[(\hat z_j(x) - z_j(x))^2\right]
\;\le\; \frac{1}{d}\,\E_\omega\!\left[\beta_j(\omega)^2\,\psi_\omega(x)^2\right]
\;\le\; \frac{K^2 R^2}{d}.
\end{equation}
Stacking the $w_j$ as the rows of $W$, summing over $j$ and exchanging the expectations over $x$ and $\omega_{1:d}$ gives
\begin{equation}
\E_{\omega_{1:d}}\!\left[\E_x\!\left[\norm{W f_\mathrm{rand}(x) - z(x)}^2\right]\right] \;\le\; \frac{k\,K^2 R^2}{d}.
\end{equation}
By Markov's inequality, the inner expectation exceeds $\epsilon$ with probability at most $k K^2 R^2/(d\,\epsilon)$, which is at most $\delta$ for $d \ge d_{\epsilon,\delta}$. \hfill $\square$

\paragraph{Remarks.} When the $\beta_j$ are bounded, a concentration inequality improves the dependence on $\delta$ to $\log(1/\delta)$ \citep{rahimi2008weighted}, and \citet{bach2017equivalence} and \citet{rudi2017generalization} give sharper rates under spectral assumptions on $\kappa$. For our purposes the content is qualitative: for targets in $\Hc_\kappa$, the number of random features needed for $\epsilon$-accurate linear decoding is finite, and grows linearly in the number of latents and in their squared RKHS norm.

\subsection{Proposition~\ref{prop:reverse} (Reverse-Predictivity Asymmetry)}

\paragraph{Statement (restated).} Suppose $f(x) = A z(x) + \mu + \varepsilon(x)$ with $\E[\varepsilon(x)\,z(x)^\top] = 0$. Then the optimal reverse probe achieves
\begin{equation}
\label{eq:rev-formula}
R^2_\mathrm{rev}
\;=\;
\frac{\tr(A\,\Cov(z)\,A^\top)}{\tr(A\,\Cov(z)\,A^\top) + \tr\Cov(\varepsilon)}.
\end{equation}
This quantity can be arbitrarily small even when forward $R^2 = 1$.

\paragraph{Proof of~\eqref{eq:rev-formula}.} Without loss of generality assume $\E[z] = 0$; the constant $\mu$ is absorbed into the intercept of the reverse probe. The least-squares reverse probe is
\begin{equation}
W_\mathrm{rev} \;=\; \E[f\,z^\top]\,\Cov(z)^{+},
\end{equation}
where $^{+}$ denotes the Moore--Penrose pseudoinverse. The pseudoinverse matters in practice: on SVG-World twelve of the 32 stored latent coordinates are constant, so $\Cov(z)$ is singular. Substituting $f = Az + \varepsilon$ and using $\E[\varepsilon\,z^\top] = 0$ gives $\E[f\,z^\top] = A\,\Cov(z)$, so $W_\mathrm{rev}\,z = A\,\Cov(z)\Cov(z)^{+} z = A\,P z$, with $P$ the orthogonal projector onto the range of $\Cov(z)$. Since centered $z$ lies in that range almost surely, $Pz = z$, so $W_\mathrm{rev}\,z = A z$ and the residual is $f - W_\mathrm{rev}\,z = \varepsilon$. Hence
\begin{equation}
\E\!\left[\norm{f - W_\mathrm{rev}\,z}^2\right] \;=\; \tr\Cov(\varepsilon).
\end{equation}
For the denominator, the cross term vanishes by $\E[\varepsilon\,z^\top] = 0$, so
\begin{equation}
\E\!\left[\norm{f - \E f}^2\right] \;=\; \tr(A\,\Cov(z)\,A^\top) + \tr\Cov(\varepsilon).
\end{equation}
Combining,
\begin{equation}
R^2_\mathrm{rev}
\;=\; 1 - \frac{\tr\Cov(\varepsilon)}{\tr(A\,\Cov(z)\,A^\top) + \tr\Cov(\varepsilon)}
\;=\; \frac{\tr(A\,\Cov(z)\,A^\top)}{\tr(A\,\Cov(z)\,A^\top) + \tr\Cov(\varepsilon)}. \tag*{$\square$}
\end{equation}

\paragraph{Asymmetry between forward and reverse $R^2$.} Let $z \in \R^k$ with $\Cov(z) = I_k$, and let $\eta(x) \in \R^M$ satisfy $\Cov(\eta) = s^2 I_M$ and $\E[\eta\,z^\top] = 0$, for example a nonlinear lift that is uncorrelated with $z$. Define $f(x) = (z(x)^\top, \eta(x)^\top)^\top \in \R^{k+M}$. Then $f = Az + \varepsilon$ with $A = (I_k\;\;0)^\top$ and $\varepsilon = (0,\eta^\top)^\top$. The forward probe $(I_k\;\;0)$ recovers $z$ exactly, so forward $R^2 = 1$, while~\eqref{eq:rev-formula} gives
\begin{equation}
R^2_\mathrm{rev} \;=\; \frac{k}{k + s^2 M} \;\longrightarrow\; 0 \qquad (M \to \infty).
\end{equation}
This is the kernel-machine regime: $z$ is exactly linearly recoverable but carries a vanishing share of the variance of $f$. \hfill $\square$

\paragraph{Remark.} In the simulation of \S\ref{sec:simulation}, the network trained with weight decay reaches $R^2_\mathrm{rev}$ close to $1$ and the unregularized one does not, consistent with the hypothesis that penalizing the readout suppresses variance of the $\eta$ type. Pretrained encoders lie between the two regimes (\S\S\ref{sec:reverse}--\ref{sec:breadth}).

\subsection{Proposition~\ref{prop:ceiling} (Ceiling)}

\paragraph{Statement (restated).} For any $f$, let $\lambda_1 \ge \dots \ge \lambda_d$ be the eigenvalues of $\Cov(f)$ and $r = \rank\Cov(z)$. Then
\begin{equation}
\label{eq:ceiling}
R^2_\mathrm{rev} \;\le\; C_r \;=\; \frac{\sum_{i \le r} \lambda_i}{\sum_{i \le d} \lambda_i},
\end{equation}
with equality iff the range of $W_\mathrm{rev}$ lies in a leading $r$-dimensional principal subspace of $f$ and the reverse-probe residual $f - W_\mathrm{rev} z$ has no variance in that range.

\paragraph{Proof of~\eqref{eq:ceiling}.} Without loss of generality assume $\E[z] = 0$ and $\E[f] = 0$. Let $\hat f = W_\mathrm{rev}\,z$ be the optimal linear predictor of $f$ from $z$ and $e = f - \hat f$ the residual. The normal equations give $\E[e\,z^\top] = 0$, hence $\E[e\,\hat f^\top] = 0$ and
\begin{equation}
\Sigma \;:=\; \Cov(f) \;=\; S + \Cov(e),
\qquad
S \;:=\; \Cov(\hat f) \;=\; W_\mathrm{rev}\,\Cov(z)\,W_\mathrm{rev}^\top.
\end{equation}
Both terms are positive semidefinite, so $S \preceq \Sigma$, and $\rank S \le \rank\Cov(z) = r$. As in the proof of~\eqref{eq:rev-formula},
\begin{equation}
R^2_\mathrm{rev} \;=\; 1 - \frac{\tr\Cov(e)}{\tr\Sigma} \;=\; \frac{\tr S}{\tr\Sigma}.
\end{equation}
Let $P$ be the orthogonal projector onto $\operatorname{range}(S)$, of rank at most $r$. Since $PSP = S$ and $S \preceq \Sigma$,
\begin{equation}
\tr S \;=\; \tr(PSP) \;\le\; \tr(P\,\Sigma\,P) \;\le\; \sum_{i \le r} \lambda_i,
\end{equation}
where the last step is Ky Fan's maximum principle: over orthogonal projectors of rank at most $r$, $\tr(P\,\Sigma\,P)$ is maximized by a projector onto a leading $r$-dimensional eigenspace of $\Sigma$. Dividing by $\tr\Sigma = \sum_{i \le d}\lambda_i$ gives $R^2_\mathrm{rev} \le C_r$.

Equality in the first inequality holds iff $\tr(P\,\Cov(e)\,P) = 0$, that is, iff the residual has no variance in $\operatorname{range}(S)$. Equality in the second holds iff this range lies in a leading $r$-dimensional principal subspace of $\Sigma$ (any such subspace, if eigenvalues are tied). \hfill $\square$

\paragraph{Remark.} The bound depends on $f$ only through its spectrum, so $R^2_\mathrm{rev}/C_r$ is the fraction of the variance attainable by a rank-$r$ linear image of the latents that the specified factors actually take. For a scalar factor, $C_1 = \lambda_1/\tr\Sigma$, so even a perfectly isolated factor cannot exceed the top-eigenvalue share. The ceiling is generally not attained, so the ratio normalizes against a bound rather than an achievable value, and we report it next to raw $R^2_\mathrm{rev}$, not instead of it. It is a population statement; held-out estimates of the ratio can exceed $1$ by sampling error.

% ============================================================================
\section{Simulation Details}
\label{app:simulation}

\subsection{Specification}

Latents $z \sim \mathrm{Unif}([-1,1]^2)$ are lifted by a frozen random Fourier map, $x_j = \cos(w_j^\top z + b_j)$ for $j = 1, \dots, 256$, with $w_j \sim \mathcal N(0, 3^2 I_2)$ and $b_j \sim \mathrm{Unif}[0, 2\pi]$ drawn once from a fixed seed and never trained; a linear map of $z$ explains $0.2452$ of the variance of this lift. The network is $f(x) = W_2\,\mathrm{ReLU}(W_1 x + b_1) + b_2$ with $W_1 \in \R^{h \times 256}$ and $W_2 \in \R^{h \times h}$, so $h$ is both the hidden and the feature width, with PyTorch's default initialization. It is trained jointly with a linear readout $\hat z = U f$ under mean-squared error with AdamW (learning rate $2\cdot 10^{-3}$, 2000 steps, batch 256, weight decay $\lambda \in \{0, 1\}$ on all parameters). We use $8{,}000$ training and $2{,}000$ test points and widths $h \in \{4, 8, \dots, 1024\}$. Probes are ridge regressions with penalty $10^{-3}$; for OOD $R^2$ the probe is trained on $\norm{z} < 0.8$ and tested on $\norm{z} \ge 0.8$. Fig.~\ref{fig:simulation-perf} averages 3 seeds; Table~\ref{tab:sim-1024} and the audit below use 5.

\subsection{Which Spectral Statistic Fig.~\ref{fig:simulation-perf} Shows}

The spectral-concentration panel of Fig.~\ref{fig:simulation-perf} plots the share of total feature variance that lies in the top $r = 2$ principal components of $f$ and is linear in $z$, computed without whitening. This is not the whitened statistic SC defined in \S\ref{sec:diagnostics}, and the two behave differently here (Table~\ref{tab:sim-1024}). Under the whitened definition both trained networks stay close to $1$ at every width, so the dip of the unregularized network disappears, while random initialization rises with width, from $0.16$ at $h = 4$ to $0.31$ at $h = 64$ and $0.71$ at $h = 1024$. In the simulation, whitened SC therefore separates trained from random networks clearly at small and moderate widths but only partly at $h = 1024$; on SVG-World and CelebA it separates every pretrained transformer from its random initialization by a factor of five or more (Tables~\ref{tab:encoders} and~\ref{tab:celeba}). Fig.~\ref{fig:simulation-spectrum}(d) projects onto the top $m$ PCs without whitening.

\begin{table}[h]
\centering\footnotesize\setlength{\tabcolsep}{5pt}
\caption{\textbf{Simulation diagnostics at $h = 1024$} (mean over 5 seeds). Unwhitened top-2: share of total variance in the top two PCs that is linear in $z$, the statistic plotted as spectral concentration in Fig.~\ref{fig:simulation-perf}. SC: whitened definition of \S\ref{sec:diagnostics}. This run is independent of the one reported in \S\ref{sec:simulation}; reverse $R^2$ differs by at most $0.003$.}
\label{tab:sim-1024}
\begin{tabular}{lccccccc}
\toprule
Condition & Fwd $R^2$ & $R^2_\mathrm{rev}$ & $C_2$ & Unwhitened top-2 & SC & OOD $R^2$ \\
\midrule
Trained, $\lambda = 1$ & 1.000 & 0.996 & 0.996 & 0.995 & 1.000 & 1.000 \\
Trained, $\lambda = 0$ & 1.000 & 0.786 & 0.797 & 0.781 & 0.983 & 0.996 \\
Random init & 1.000 & 0.213 & 0.258 & 0.181 & 0.713 & 0.892 \\
\bottomrule
\end{tabular}
\end{table}

\subsection{Non-Degeneracy Audit}

At $h = 1024$ (5 seeds, 2000 held-out points), the network trained with $\lambda = 1$ visits $1988.2$ distinct ReLU activation regions on average (minimum $1983$); $93.67\%$ of units are active on between $5\%$ and $95\%$ of inputs; the dead fraction is $0.0359$ and the mean active fraction $0.2683$; and the input--output Jacobian has a coefficient of variation of $0.6573$ across test points. An affine network would have one region and a constant Jacobian; no seed at any width did. As a calibration, forcing every pre-activation to be positive, which makes the network affine on its data, gives $1$ region, active fraction $1.000$ and Jacobian coefficient of variation $2.4 \times 10^{-15}$, so the audit detects the failure it is meant to rule out. High reverse $R^2$ alone would not establish nonlinearity: an affine map that projects $x$ onto the two directions of its latent-predictable component reaches $R^2_\mathrm{rev} = 0.8992$ (trained network in the same run: $0.9936$), so the argument rests on the activation statistics above.

\paragraph{Spectrum.} On a linear scale, the top two eigenvalues of $\Cov(f)$ carry $99.57\%$ of the variance ($0.5657$ and $0.4300$, against $0.1384$ and $0.1151$ at random initialization). Later eigenvalues are small but nonzero, so the covariance is nearly, not exactly, rank two.

% ============================================================================
\section{SVG-World Dataset Details}
\label{app:svg-world}

Table~\ref{tab:svg-card} summarizes the dataset. The latent vector has 32 dimensions, of which 20 are active: for each of the four people, an $(x, z)$ position, a facing direction and a skin hue; for each of the two animals, an $(x, z)$ position. Facing is discretized inside the renderer as $\lfloor 3.999\,z \rfloor$, so it takes four values and its $R^2$ is bounded below $1$ even for a perfect probe. The remaining 12 dimensions are shirt colors fixed by the renderer and are excluded from all probes.

\begin{table}[h]
\centering\small
\caption{SVG-World dataset card.}
\label{tab:svg-card}
\begin{tabular}{lp{0.72\linewidth}}
\toprule
Field & Value \\
\midrule
Latent vector & $z \in [0,1]^{32}$; 20 active, 12 fixed by the renderer \\
Person $\times 4$ & $(x, z)$ position in a fixed lane; facing $\in \{$S, E, N, W$\}$; skin hue $0.05 + 0.06\,z$ (HSV) \\
Animal $\times 2$ & $(x, z)$ position in a fixed lane (AquaWorld: dolphin, turtle; WestWorld: brown and white horse) \\
Fixed (12 dims) & shirt HSV per person; excluded from all probes \\
Sampling & $z \sim \mathcal U[0,1]^{32}$ i.i.d., seed 42 \\
Occlusion & objects drawn back to front, so they may overlap \\
Rendering & SVG $\to$ \texttt{cairosvg}; one layout, two styles \\
Rasterization & $600 \times 420$ SVG padded to $600 \times 600$, resized to $224 \times 224$ RGB \\
Splits & AquaWorld, WestWorld; $N = 10{,}000$ each; 50/50 train/test \\
Probed in the main text & the 12 position latents \\
\bottomrule
\end{tabular}
\end{table}

% ============================================================================
\section{Experimental Details}
\label{app:experiments}

This appendix gives the encoders, the probing protocols, compute, and the full results behind \S\ref{sec:breadth}.

\paragraph{Encoders.} All checkpoints are HuggingFace models pinned to a fixed revision (Table~\ref{tab:checkpoints}). We take one pooled vector from the final block, before any task or projection head. Each random control is the same architecture built from its configuration with default initialization; DINOv3-B/16 uses seeds 42--46 and the other encoders seed 42 unless stated otherwise. Inputs are $224 \times 224$, preprocessed as specified in each model's pinned preprocessor configuration. For ResNet-50, the random control keeps its untrained BatchNorm running statistics; calibrating them on the evaluation images would no longer be a data-free control, so we report the uncalibrated control and keep ResNet-50 out of all counts.

\begin{table}[h]
\centering\footnotesize\setlength{\tabcolsep}{4pt}
\caption{Encoders, checkpoints and readouts.}
\label{tab:checkpoints}
\resizebox{\linewidth}{!}{%
\begin{tabular}{lll}
\toprule
Encoder & Checkpoint & Readout ($d$) \\
\midrule
DINOv3-S/16 & \texttt{facebook/dinov3-vits16-pretrain-lvd1689m} & CLS (384) \\
DINOv3-B/16 & \texttt{facebook/dinov3-vitb16-pretrain-lvd1689m} & CLS or mean patch (768) \\
DINOv3-L/16 & \texttt{facebook/dinov3-vitl16-pretrain-lvd1689m} & CLS (1024) \\
DINOv2-B/14 & \texttt{facebook/dinov2-base} & CLS (768) \\
DeiT-B/16 & \texttt{facebook/deit-base-patch16-224} & CLS (768) \\
CLIP-B/16 & \texttt{openai/clip-vit-base-patch16} & pooled CLS, before projection (768) \\
ResNet-50 & \texttt{microsoft/resnet-50} & global average pool (2048) \\
\bottomrule
\end{tabular}}
\end{table}

\paragraph{Probes.} All probes are ridge regressions with an intercept. Two protocols are used. The DINOv3-B/16 analyses of \S\S\ref{sec:dino-forward}--\ref{sec:ood} (Table~\ref{tab:dino-forward}, Figs.~\ref{fig:forward-random}, \ref{fig:reverse} and~\ref{fig:ood}, and App.~\ref{app:reverse-ablations}) use a fixed penalty $\alpha = 1$ and one 50/50 split (seed 42) with the same indices in both worlds. All intervals in \S\ref{sec:reverse} and all results in \S\ref{sec:breadth} use the revised protocol: $\alpha$ chosen by 5-fold cross-validation inside the training half over $[10^{-3}, 10^{5}]$, five repeated 50/50 splits over scene identifiers (the two styles share each split), float64, and held-out reverse $R^2 = 1 - \sum_j \norm{f_j - \hat f_j}^2 / \sum_j \norm{f_j - \bar f_j}^2$ on the test half. Point estimates are on the shared split; intervals are paired bootstraps over held-out units with 10{,}000 resamples, made hierarchical over units and control initializations where there are five controls. For DINOv3-B/16 on SVG-World, moving from the first protocol to the second changes $R^2_\mathrm{rev}$ by at most $0.0042$ in every condition. It does change forward $R^2$ for random initializations: their features have 14 to 29 times less total variance than pretrained features, so a fixed $\alpha = 1$ regularizes them much more strongly. With cross-validated $\alpha$, random-init DINOv3-B/16 reaches forward $R^2 = 0.66$ on CLS features (Table~\ref{tab:encoders}), against $0.33$ under the fixed penalty (Table~\ref{tab:dino-forward}); pretrained forward $R^2$ changes little ($0.91$ against $0.88$). Cross-style probes are fitted on the training half of the source world and evaluated on the test half of the target world, with features centered on each world's own mean. For spectral concentration, PCA whitening is fitted on training rows only.

\paragraph{Factor encodings.} On dSprites, 3D Shapes and MPI3D-realistic we draw 20{,}000 factor-balanced images (each factor's marginal is uniform, every factor combination appears at most once; sample seed 1234, with seeds 7 and 2024 as replicates) and split them 50/50 by factor combination. The main results use the \emph{canonical} encoding: ordinal factors min-max scaled to $[0,1]$, cyclic factors as a $(\sin, \cos)$ pair over the period of each dataset's stored angle grid, categorical factors excluded. Two alternatives test sensitivity to this choice: \emph{ordinal}, every factor as its min-max-scaled index with no trigonometric encoding, and \emph{with one-hot}, the canonical block plus one-hot columns for the categorical factors. The factor rank $r$ is the rank of the encoded target block on the training half. The checkerboard test holds out half of the joint combinations of the two factors with the most values (those with odd index sum), so every marginal value is seen in training but half of the combinations are not.

\paragraph{Compute.} Feature extraction runs on a single NVIDIA H100 GPU and dominates the cost: DINOv3-B/16 processes about 1{,}100 images per second and DINOv3-L/16 about 330, so a pretrained and random pair on 20{,}000 images takes under three minutes. The full CelebA run takes 20--30 minutes; the simulation and all probes run on CPU. % check: confirm H100 for all lanes

\subsection{Encoders on SVG-World}

Table~\ref{tab:dino-forward} gives the per-latent forward decoding behind \S\ref{sec:dino-forward} and Fig.~\ref{fig:forward-random} (fixed-penalty protocol).

\begin{table}[ht]
\centering
\caption{\textbf{DINOv3-B/16 Forward Decoding on SVG-World, per-dim within-domain.} Pearson $\rho$ and $R^2$ from a ridge probe trained on each world separately, averaged over AquaWorld and WestWorld. \emph{Pre}: pretrained. \emph{Rand}: freshly initialized, no training. The pretraining contribution is clearest on small foreground objects (animal $\Delta\rho = 0.58$ on CLS, $0.68$ on patches). Position latents only; fixed ridge penalty $\alpha = 1$.}
\label{tab:dino-forward}
% \small
\setlength{\tabcolsep}{4pt}
\begin{tabular}{l rrrr rrrr}
% \toprule
 & \multicolumn{4}{c}{CLS token} & \multicolumn{4}{c}{Mean-pooled patches} \\
\cmidrule(lr){2-5}\cmidrule(lr){6-9}
 & \multicolumn{2}{c}{$\rho$} & \multicolumn{2}{c}{$R^2$} & \multicolumn{2}{c}{$\rho$} & \multicolumn{2}{c}{$R^2$} \\
\cmidrule(lr){2-3}\cmidrule(lr){4-5}\cmidrule(lr){6-7}\cmidrule(lr){8-9}
latent & pre & rand & pre & rand & pre & rand & pre & rand \\
\midrule
\texttt{p0\_x} & 0.97 & 0.71 & 0.94 & 0.47 & 0.93 & 0.68 & 0.85 & 0.41 \\
\texttt{p0\_z} & 0.95 & 0.80 & 0.89 & 0.61 & 0.90 & 0.76 & 0.80 & 0.55 \\
\texttt{p1\_x} & 0.91 & 0.49 & 0.82 & 0.21 & 0.82 & 0.44 & 0.65 & 0.16 \\
\texttt{p1\_z} & 0.95 & 0.72 & 0.89 & 0.50 & 0.90 & 0.70 & 0.80 & 0.45 \\
\texttt{p2\_x} & 0.93 & 0.50 & 0.86 & 0.20 & 0.86 & 0.46 & 0.71 & 0.16 \\
\texttt{p2\_z} & 0.95 & 0.60 & 0.90 & 0.32 & 0.90 & 0.58 & 0.79 & 0.25 \\
\texttt{p3\_x} & 0.95 & 0.68 & 0.91 & 0.42 & 0.91 & 0.63 & 0.81 & 0.30 \\
\texttt{p3\_z} & 0.93 & 0.74 & 0.87 & 0.52 & 0.88 & 0.73 & 0.76 & 0.42 \\
\midrule
\texttt{dolphin\_x} & 0.93 & 0.42 & 0.86 & 0.25 & 0.95 & 0.24 & 0.89 & 0.07 \\
\texttt{dolphin\_z} & 0.97 & 0.50 & 0.94 & 0.31 & 0.98 & 0.45 & 0.96 & 0.25 \\
\texttt{turtle\_x} & 0.88 & 0.18 & 0.76 & 0.03 & 0.94 & 0.13 & 0.86 & 0.02 \\
\texttt{turtle\_z} & 0.95 & 0.32 & 0.90 & 0.09 & 0.97 & 0.29 & 0.94 & 0.06 \\
\midrule
\textbf{person mean} & \textbf{0.94} & \textbf{0.65} & \textbf{0.89} & \textbf{0.41} & \textbf{0.89} & \textbf{0.62} & \textbf{0.77} & \textbf{0.34} \\
\textbf{animal mean} & \textbf{0.93} & \textbf{0.35} & \textbf{0.87} & \textbf{0.17} & \textbf{0.96} & \textbf{0.28} & \textbf{0.91} & \textbf{0.10} \\
\textbf{all mean} & \textbf{0.94} & \textbf{0.55} & \textbf{0.88} & \textbf{0.33} & \textbf{0.91} & \textbf{0.51} & \textbf{0.82} & \textbf{0.26} \\
\bottomrule
\end{tabular}
\end{table}

Table~\ref{tab:encoders} reports the seven-encoder panel behind \S\ref{sec:encoders}, under the revised protocol and averaged over the two styles. Every transformer has a reverse-predictivity gap whose interval excludes zero, and for every transformer pretraining raises spectral concentration by a factor of five or more. The cross-style correlation separates most encoders from their controls but not all: CLIP's random initialization transfers almost as well as the pretrained model ($0.31$ against $0.34$), and ResNet-50's transfers better. On AquaWorld CLS features, pretrained DINOv3-B/16 needs the top 17 principal components to contain $90\%$ of its latent-predictable variance, against 55--59 for its five random initializations.

\begin{table}[ht]
\centering\footnotesize\setlength{\tabcolsep}{3.2pt}
\caption{\textbf{Encoders on SVG-World} (12 position latents, $r = 12$; mean over the two styles; random control seed 42). Gap: $R^2_\mathrm{rev}$ pretrained minus random, with 95\% paired bootstrap interval. SC: spectral concentration $R^2_\mathrm{rev}(r)$. $\rho_\times$: mean cross-style correlation over both transfer directions. Each cell: pretrained / random.}
\label{tab:encoders}
\resizebox{\linewidth}{!}{%
\begin{tabular}{lcccccc}
\toprule
Encoder (readout) & Fwd $R^2$ & $R^2_\mathrm{rev}$ & Gap & $R^2_\mathrm{rev}/C_r$ & SC & $\rho_\times$ \\
\midrule
DINOv3-S/16 (CLS) & 0.86 / 0.45 & 0.257 / 0.023 & 0.235 [0.229, 0.242] & 0.38 / 0.04 & 0.27 / 0.03 & 0.47 / 0.09 \\
DINOv3-B/16 (CLS) & 0.91 / 0.66 & 0.188 / 0.023 & 0.164 [0.160, 0.169] & 0.31 / 0.04 & 0.26 / 0.03 & 0.46 / 0.10 \\
DINOv3-B/16 (patches) & 0.93 / 0.59 & 0.284 / 0.024 & 0.259 [0.253, 0.264] & 0.47 / 0.05 & -- & 0.57 / 0.15 \\
DINOv3-L/16 (CLS) & 0.91 / 0.74 & 0.169 / 0.024 & 0.145 [0.139, 0.151] & 0.26 / 0.05 & 0.24 / 0.02 & 0.48 / 0.06 \\
DINOv2-B/14 (CLS) & 0.91 / 0.58 & 0.224 / 0.028 & 0.196 [0.190, 0.202] & 0.35 / 0.05 & 0.26 / 0.04 & 0.34 / 0.07 \\
DeiT-B/16 (CLS) & 0.89 / 0.55 & 0.186 / 0.020 & 0.165 [0.159, 0.170] & 0.27 / 0.04 & 0.23 / 0.02 & 0.26 / 0.02 \\
CLIP-B/16 (pooled) & 0.87 / 0.45 & 0.174 / 0.020 & 0.153 [0.148, 0.158] & 0.27 / 0.04 & 0.21 / 0.02 & 0.34 / 0.31 \\
\midrule
ResNet-50 (pool) & 0.93 / 0.95 & 0.325 / 0.220 & 0.105 [0.100, 0.111] & 0.55 / 0.49 & 0.40 / 0.35 & 0.27 / 0.46 \\
\bottomrule
\end{tabular}}
\end{table}

\subsection{Controlled Benchmarks}

Table~\ref{tab:benchmarks} gives the full results behind \S\ref{sec:benchmarks}. For DINOv3-B/16 on dSprites and 3D Shapes, the random initialization has a much higher $R^2_\mathrm{rev}$ than the pretrained encoder, and normalizing by the ceiling does not change this, because the two ceilings are similar ($0.81$ and $0.82$ on dSprites). These random networks are nearly one-dimensional (participation ratio $1.9$ and $3.8$), and most of that variance is linear in the factors. On renders of a single object against a plain background, simple factors such as position, scale and color account for much of the pixel variance: on raw $32 \times 32$ pixels, $R^2_\mathrm{rev}$ is $0.84$ on 3D Shapes, against $0.17$ on SVG-World (Table~\ref{tab:pixels}). A random encoder that stays close to its input can therefore score highly on D1 without having learned anything, which is why the diagnostics are necessary rather than sufficient conditions. On MPI3D-realistic the ceilings differ (pretrained $0.35$, random $0.96$), and after normalization the pretrained encoder is ahead ($0.53$ against $0.42$). CLIP's random initialization on dSprites is not low-dimensional (participation ratio $18.7$), and there the pretrained encoder is ahead on both raw and normalized $R^2_\mathrm{rev}$. The checkerboard test follows forward $R^2$ closely: pretrained ahead on dSprites and MPI3D-realistic, tied on 3D Shapes.

\begin{table}[h]
\centering\footnotesize\setlength{\tabcolsep}{3.5pt}
\caption{\textbf{Controlled benchmarks} (canonical encoding, sample seed 1234; pretrained / mean of five random initializations). D3: $R^2$ on held-out checkerboard combinations, against random seed 42 only. ResNet-50 uses the data-free control and is reported separately.}
\label{tab:benchmarks}
\begin{tabular}{llccccccc}
\toprule
Dataset ($r$) & Encoder & Fwd $R^2$ & $R^2_\mathrm{rev}$ & $C_r$ & $R^2_\mathrm{rev}/C_r$ & SC & PR & D3 \\
\midrule
dSprites (5) & DINOv3-B/16 & 0.42 / 0.16 & 0.09 / 0.61 & 0.81 / 0.82 & 0.11 / 0.74 & 0.08 / 0.41 & 4.4 / 1.9 & 0.42 / 0.17 \\
 & CLIP-B/16 & 0.30 / 0.04 & 0.17 / 0.09 & 0.67 / 0.46 & 0.25 / 0.20 & 0.12 / 0.14 & 6.1 / 18.7 & 0.31 / 0.04 \\
3D Shapes (8) & DINOv3-B/16 & 0.96 / 0.96 & 0.11 / 0.67 & 0.70 / 0.92 & 0.15 / 0.72 & 0.26 / 0.24 & 8.7 / 3.8 & 0.95 / 0.96 \\
 & CLIP-B/16 & 0.96 / 0.97 & 0.15 / 0.90 & 0.61 / 1.00 & 0.25 / 0.91 & 0.17 / 0.32 & 16.6 / 2.3 & 0.95 / 0.97 \\
MPI3D-real.\ (4) & DINOv3-B/16 & 0.85 / 0.61 & 0.19 / 0.40 & 0.35 / 0.96 & 0.53 / 0.42 & 0.26 / 0.27 & 19.8 / 3.1 & 0.85 / 0.61 \\
 & CLIP-B/16 & 0.76 / 0.51 & 0.14 / 0.09 & 0.35 / 0.97 & 0.41 / 0.09 & 0.22 / 0.19 & 20.8 / 2.6 & 0.77 / 0.51 \\
\midrule
dSprites & ResNet-50 & 0.24 / 0.19 & 0.25 / 0.54 & 0.71 / 0.76 & 0.35 / 0.72 & 0.25 / 0.46 & 6.7 / 3.1 & 0.24 / 0.19 \\
3D Shapes & ResNet-50 & 0.88 / 0.99 & 0.08 / 0.39 & 0.62 / 0.97 & 0.13 / 0.40 & 0.19 / 0.39 & 13.3 / 1.2 & 0.87 / 0.99 \\
MPI3D-real. & ResNet-50 & 0.79 / 0.68 & 0.20 / 0.86 & 0.37 / 0.99 & 0.53 / 0.87 & 0.30 / 0.29 & 19.2 / 1.1 & 0.79 / 0.67 \\
\bottomrule
\end{tabular}
\end{table}

The $R^2_\mathrm{rev}$ gaps are stable across control initializations (hierarchical intervals exclude zero in every cell) and across dataset samples (no gap moves by more than $0.0204$ over three samples), but not across factor encodings (Table~\ref{tab:encodings}). Of the six transformer--benchmark pairs, four keep the sign of the gap under all three encodings; DINOv3-B/16 on 3D Shapes and CLIP-B/16 on MPI3D-realistic do not. A reverse-predictivity comparison is therefore defined only for a fully specified factor set and its encoding, and should not be read as a score of encoder quality.

\begin{table}[h]
\centering\footnotesize\setlength{\tabcolsep}{5pt}
\caption{\textbf{Sensitivity to the factor encoding.} $R^2_\mathrm{rev}$ gap (pretrained minus mean of five random initializations), sample seed 1234.}
\label{tab:encodings}
\begin{tabular}{llccc}
\toprule
Dataset & Encoder & Canonical & Ordinal & With one-hot \\
\midrule
dSprites & DINOv3-B/16 & $-0.516$ & $-0.398$ & $-0.287$ \\
 & CLIP-B/16 & $+0.077$ & $+0.097$ & $+0.128$ \\
3D Shapes & DINOv3-B/16 & $-0.559$ & $+0.075$ & $-0.094$ \\
 & CLIP-B/16 & $-0.754$ & $-0.073$ & $-0.492$ \\
MPI3D-realistic & DINOv3-B/16 & $-0.213$ & $-0.344$ & $-0.478$ \\
 & CLIP-B/16 & $+0.057$ & $-0.213$ & $-0.418$ \\
\midrule
dSprites & ResNet-50 & $-0.297$ & $-0.332$ & $-0.274$ \\
3D Shapes & ResNet-50 & $-0.316$ & $-0.010$ & $+0.023$ \\
MPI3D-realistic & ResNet-50 & $-0.661$ & $-0.639$ & $-0.586$ \\
\bottomrule
\end{tabular}
\end{table}

\begin{table}[h]
\centering\footnotesize\setlength{\tabcolsep}{6pt}
\caption{\textbf{Pixel-space control.} Diagnostics computed on raw images downsampled to $32 \times 32 \times 3$ ($d = 3072$), canonical encoding.}
\label{tab:pixels}
\begin{tabular}{lcccc}
\toprule
Dataset & Fwd $R^2$ & $R^2_\mathrm{rev}$ & $C_r$ & $R^2_\mathrm{rev}/C_r$ \\
\midrule
SVG-World (mean of styles) & 0.995 & 0.166 & 0.412 & 0.42 \\
dSprites & 0.496 & 0.213 & 0.489 & 0.44 \\
3D Shapes & 0.998 & 0.836 & 0.885 & 0.94 \\
MPI3D-realistic & 0.636 & 0.389 & 0.676 & 0.58 \\
\bottomrule
\end{tabular}
\end{table}

\subsection{CelebA}

We sample 20{,}000 images from the CelebA training partition (seed 0) and use its 40 binary attributes, coded in $\{0,1\}$, as the factor set ($r = 40$). DINOv3-B/16 uses the checkpoint above; CLIP-B/16 is the OpenAI ViT-B/16 model in \texttt{open\_clip}, whose image embedding we probe. Random controls use seeds 0--4. Probes use a fixed $\alpha = 1$ and one 50/50 split (seed 42); intervals are percentile bootstraps over held-out images with 1{,}000 resamples. Because the attributes are binary, recoding them (for example as $\pm 1$) is an affine map and leaves every reverse statistic unchanged, so we vary the feature normalization instead: pretrained exceeds random on $R^2_\mathrm{rev}$ for raw, per-dimension standardized and PCA-whitened features alike (Table~\ref{tab:celeba}).

\begin{table}[h]
\centering\footnotesize\setlength{\tabcolsep}{3.5pt}
\caption{\textbf{CelebA} (20{,}000 images; 40 binary attributes, $r = 40$). Pretrained: 95\% bootstrap interval over held-out images; random: mean $\pm$ s.d.\ over five initializations. Std.: $R^2_\mathrm{rev}$ after standardizing each feature dimension. SC: spectral concentration, $R^2_\mathrm{rev}$ on the top $r$ whitened PCs. PR: participation ratio.}
\label{tab:celeba}
\begin{tabular}{lllccccccc}
\toprule
Encoder & Features & Weights & Fwd $R^2$ & $R^2_\mathrm{rev}$ & Std. & $C_r$ & $R^2_\mathrm{rev}/C_r$ & SC & PR \\
\midrule
DINOv3-B/16 & CLS & pre & 0.391 & 0.170 [0.168, 0.173] & 0.160 & 0.606 & 0.281 & 0.153 & 48.2 \\
 & & rand & 0.069 & 0.067 $\pm$ 0.003 & 0.060 & 0.980 & 0.069 & 0.018 & 2.5 \\
DINOv3-B/16 & patches & pre & 0.412 & 0.245 [0.242, 0.248] & 0.226 & 0.680 & 0.360 & 0.174 & 32.1 \\
 & & rand & 0.114 & 0.090 $\pm$ 0.002 & 0.081 & 0.994 & 0.091 & 0.023 & 1.7 \\
CLIP-B/16 & image & pre & 0.411 & 0.192 [0.191, 0.195] & 0.152 & 0.694 & 0.277 & 0.155 & 45.6 \\
 & & rand & 0.115 & 0.106 $\pm$ 0.001 & 0.097 & 0.996 & 0.106 & 0.028 & 1.8 \\
\bottomrule
\end{tabular}
\end{table}

% ============================================================================
\section{Reverse Predictivity: Ablations}
\label{app:reverse-ablations}

Standardizing each feature dimension before fitting the reverse probe leaves the aggregate $R^2_\mathrm{rev}$ of DINOv3-B/16 on SVG-World within $0.01$ of its raw value: pretrained CLS goes from $0.181$ to $0.175$, random from $0.019$ to $0.020$. The pretrained-vs-random gap is therefore not an artifact of the scale of individual feature dimensions. The whitening curve of Fig.~\ref{fig:reverse} on CLS features, at $m \in \{32, 64, 128, 256, 768\}$, gives $R^2_\mathrm{rev}(m) = 0.174, 0.114, 0.067, 0.035, 0.009$ for the pretrained encoder and $0.023, 0.022, 0.018, 0.014, 0.006$ for random initialization. Evaluated at the factor rank $m = r = 12$, the curve gives $0.26$ pretrained against $0.03$ random (Table~\ref{tab:encoders}).

The whitened curve falls for $m > r$ for any representation with nonzero residual, because whitening lifts residual directions to unit variance however small they are; for an exact linear world model of three latents with small isotropic noise it exceeds $0.95$ at $m = 3$ and falls below $0.30$ at $m = 64$. The decline with $m$ is therefore not evidence against a linear world model. What separates encoders is the value near $m = r$ and how many components are needed to collect the latent-predictable variance.

% ============================================================================
\section{Cross-Domain Transfer: Per-Direction Results}
\label{app:ood}

Fig.~\ref{fig:ood} averages the two transfer directions (AquaWorld$\to$WestWorld and WestWorld$\to$AquaWorld). Under the revised protocol the directions differ. On DINOv3-B/16 CLS features, the mean correlation between the transferred probe's predictions and the target latents is $0.381$ in one direction and $0.533$ in the other, against $0.074$ and $0.099$ for random initialization (means over five initializations). Transfer $R^2$ without calibration is not meaningful, because feature scales differ between the two styles and per-world centering removes only the mean difference ($-58.8$ for the pretrained encoder, around $-2 \times 10^4$ at random initialization). After rescaling the probe output on half of the target world, transfer $R^2$ is $0.164$ and $0.313$ for the pretrained encoder, against $0.011$ and $0.018$ for random initialization (means over the same five initializations).

% ============================================================================
\section{Binding with Multiple Objects}
\label{app:binding}

\begin{figure}[t]
\centering
\includegraphics[width=\linewidth]{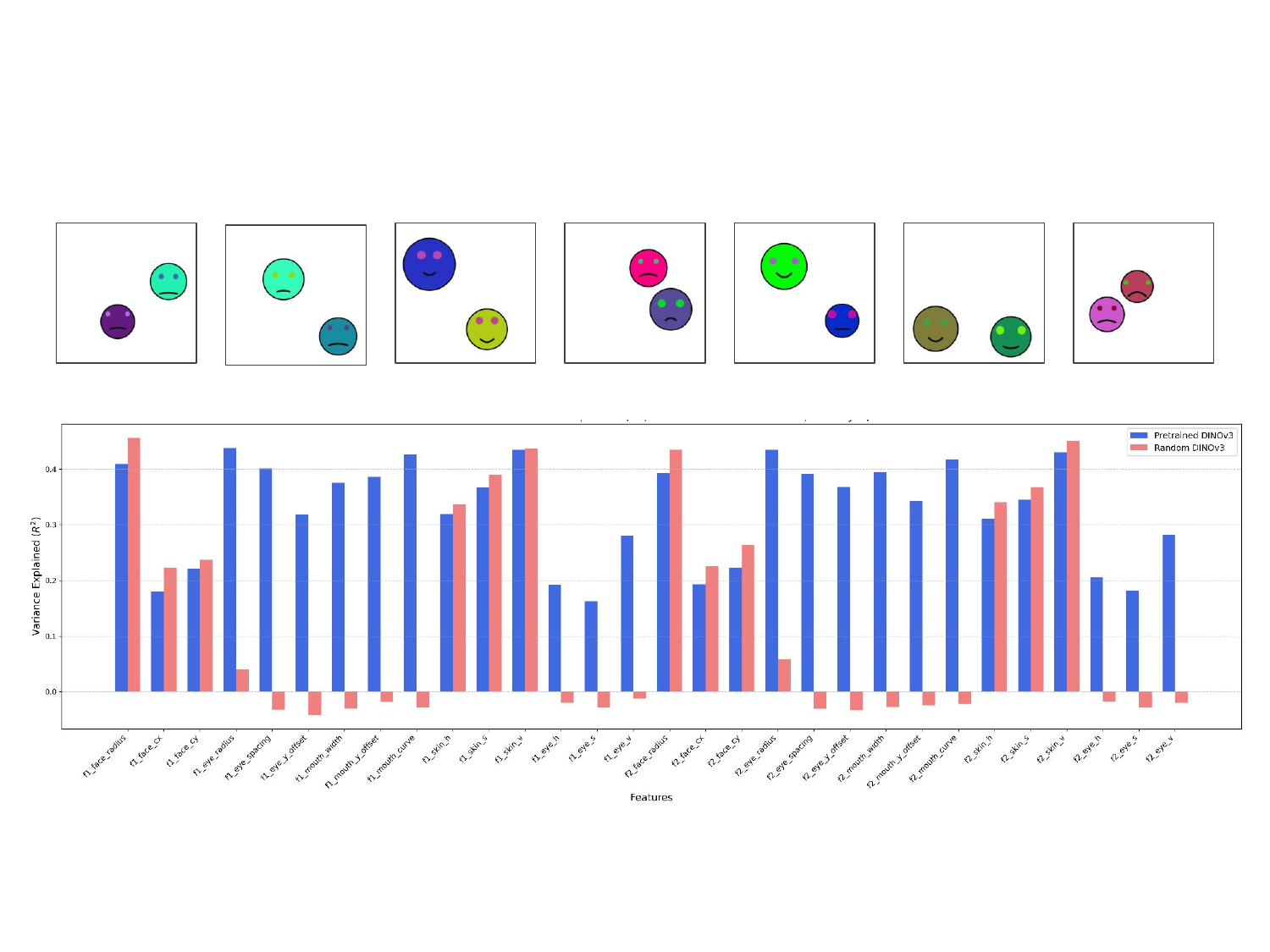}
\caption{\textbf{Two-face regime of SVG-Faces.} (a) Example renders; each scene contains two faces with independently sampled latents, ordered left to right by $x$-coordinate. (b) Per-latent test $R^2$ for pretrained and random-init DINOv3-B/16 (30 latents, 15 per face).}
\label{fig:binding}
\end{figure}

With several objects in a scene, a linear readout must keep each object's attributes paired with that object. SVG-World largely sidesteps this: each person and animal moves within its own fixed lane, so identity is given by location, and each person's position is decoded separately with $R^2 \ge 0.82$ on CLS features (Table~\ref{tab:dino-forward}). The two-face regime of the SVG-Faces precursor (App.~\ref{app:svg-faces}) tests binding directly, since two faces with independently sampled attributes appear anywhere on the canvas and are ordered left to right by their $x$-coordinate. Here decoding is far weaker than with one face (Fig.~\ref{fig:binding}): per-latent $R^2$ stays below about $0.45$ even for the pretrained encoder. Random initialization matches the pretrained encoder on face position, size and skin color, and fails on eye and mouth geometry and eye color, which pretraining recovers only partially. Neither encoder binds attributes to objects reliably under a single linear readout.

% ============================================================================
\section{SVG-Faces: A Single-Object Precursor Benchmark}
\label{app:svg-faces}

\begin{figure}[t]
\centering
\includegraphics[width=0.85\linewidth]{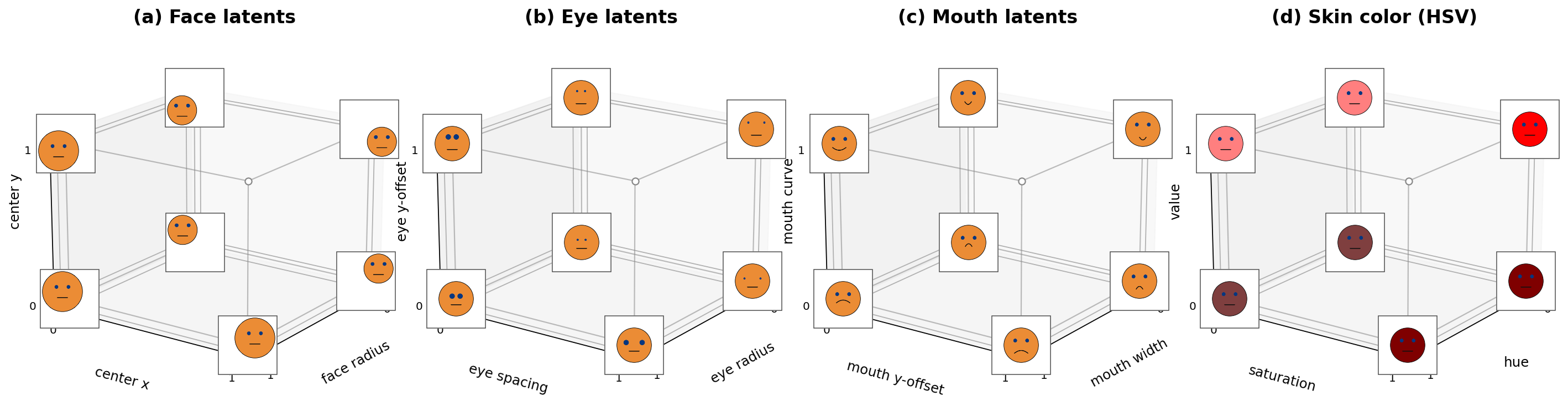}
\caption{\textbf{SVG-Faces.} Each panel varies three latents of the single-face regime while the others are held fixed: \textbf{(a)} face center $(x, y)$ and radius, \textbf{(b)} eye spacing, vertical offset and radius, \textbf{(c)} mouth vertical offset, width and curvature, and \textbf{(d)} skin hue, saturation and value. Eye color (HSV, not shown) completes the 15 latents. The renderer is deterministic.}
\label{fig:svg-faces}
\end{figure}

SVG-Faces was the first testbed for this project. Each image shows one smiley face controlled by 15 continuous latents (Fig.~\ref{fig:svg-faces}): face radius and center $(x, y)$; eye radius, spacing and vertical offset; mouth width, vertical offset and curvature; and skin and eye color in HSV. A two-face regime with 30 latents is used for the binding test of App.~\ref{app:binding}. The main results use SVG-World, which adds multiple objects and a second visual style; SVG-Faces is included as a consistency check with a different pipeline and a single object.

\paragraph{Forward decoding.} A linear probe on DINOv3-B/16 recovers all nine geometric latents with $\rho \ge 0.98$ (Fig.~\ref{fig:forward-random}). Color is harder: skin saturation and value reach $\rho \ge 0.97$, skin hue $0.93$, and eye color $0.82$--$0.91$, with eye hue lowest. The hue scatter plots bend at both ends because hue is circular and a linear probe cannot represent the wrap from $1$ back to $0$. Decoded latents re-rendered through the SVG pipeline give faithful reconstructions in distribution.

\begin{figure}[t]
\centering
\includegraphics[width=\linewidth]{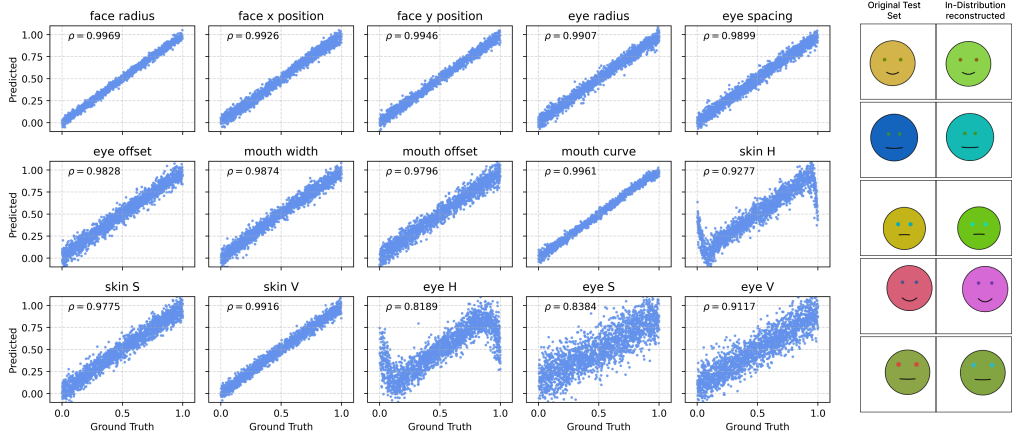}
\caption{\textbf{DINOv3-B/16 forward decoding on SVG-Faces.} (a) Predicted vs.\ ground-truth values on held-out images for the 15 single-face latents. (b) Round-trip reconstructions: latents decoded from a held-out image and re-rendered through the SVG pipeline.}
\label{fig:dino-forward-faces}
\end{figure}

\begin{figure}[t]
\centering
\includegraphics[width=0.99\linewidth]{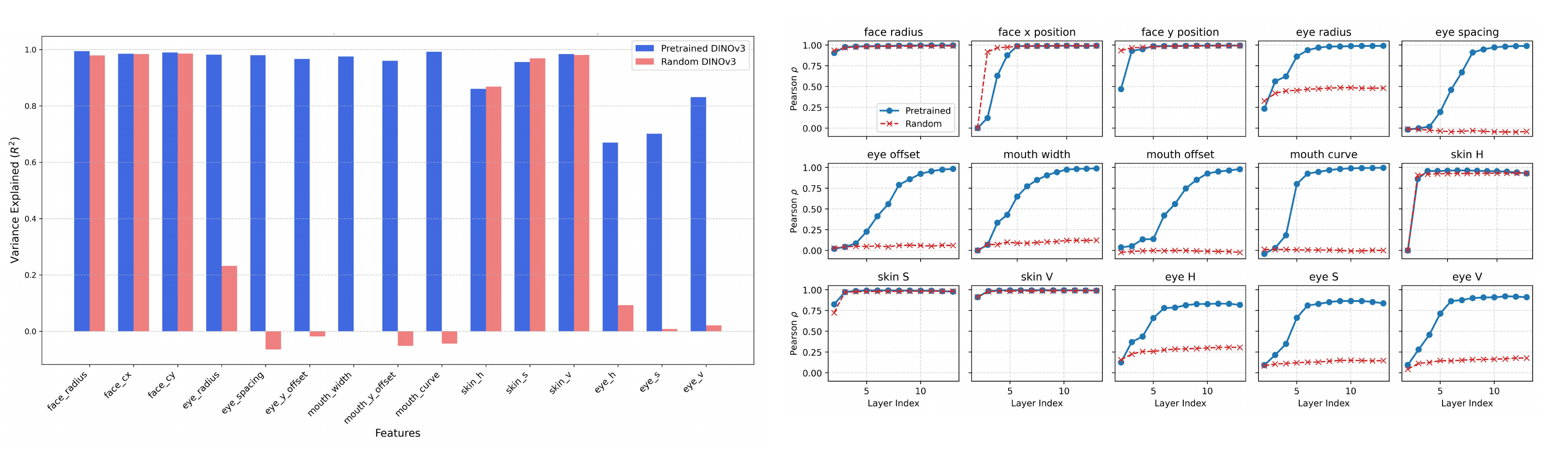}
\caption{\textbf{SVG-Faces: pretrained vs.\ random initialization.} \emph{Left:} per-latent test $R^2$ on the final block. \emph{Right:} Pearson $\rho$ per transformer block. Random initialization matches the pretrained encoder on face position, radius and skin color, and fails on eye and mouth geometry and eye color.}
\label{fig:forward-random-faces}
\end{figure}

\paragraph{Pretrained vs.\ random.} The comparison with random initialization splits the latents into two groups (Fig.~\ref{fig:forward-random-faces}). Random initialization decodes face position and radius and skin color as well as the pretrained encoder, with $R^2$ above $0.85$. It fails on the small parts, eye and mouth geometry and eye color, where its $R^2$ is at most $0.23$ and mostly near zero, while the pretrained encoder reaches $0.67$--$0.99$. Resolved by layer, the pretrained advantage on these latents builds up gradually over the transformer blocks.

\begin{figure}[t]
\centering
\includegraphics[width=\linewidth]{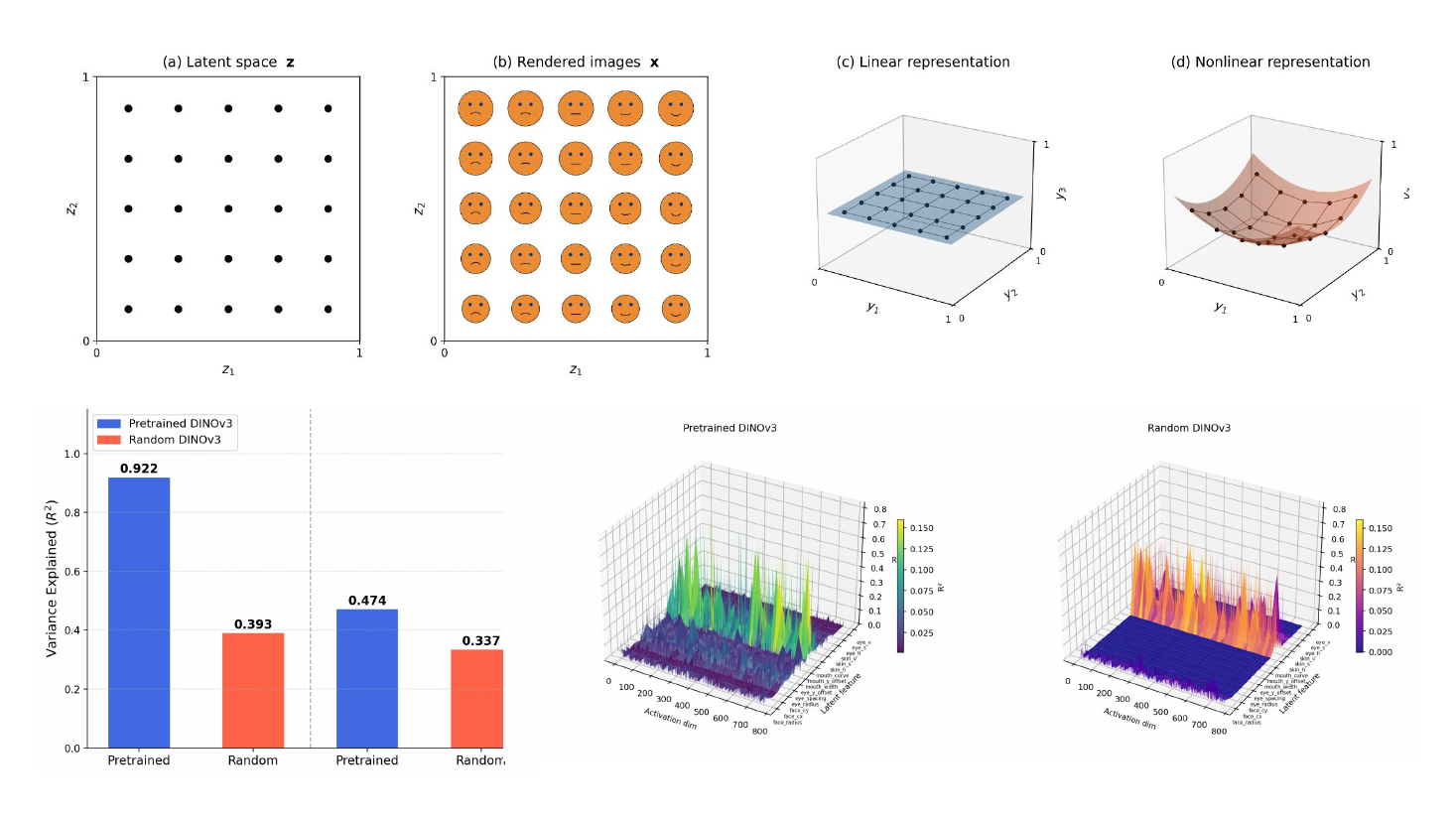}
\caption{\textbf{SVG-Faces: reverse predictivity on DINOv3-B/16.} (a) Conceptual setup. (b) Aggregate forward and reverse $R^2$ for pretrained and random initialization. (c) Contribution of each (feature dimension, latent) pair to reverse $R^2$.}
\label{fig:reverse-faces}
\end{figure}

\paragraph{Reverse predictivity.} With the unwhitened reverse probe, aggregate forward $R^2$ is $0.92$ for the pretrained encoder against $0.39$ for random initialization, and aggregate $R^2_\mathrm{rev}$ is $0.47$ against $0.34$ (Fig.~\ref{fig:reverse-faces}). The pattern differs from SVG-World: here the reverse gap is smaller than the forward gap in both absolute and relative terms, whereas on SVG-World the reverse ratio is much larger than the forward ratio (\S\ref{sec:reverse}). One plausible reason is that a single large face, whose position, size and color are among the latents, accounts for most of the pixel variation, so even random features carry a large latent-aligned share, as on the controlled benchmarks (Table~\ref{tab:pixels}). We have not tested this explanation.

\begin{figure}[t]
\centering
\includegraphics[width=\linewidth]{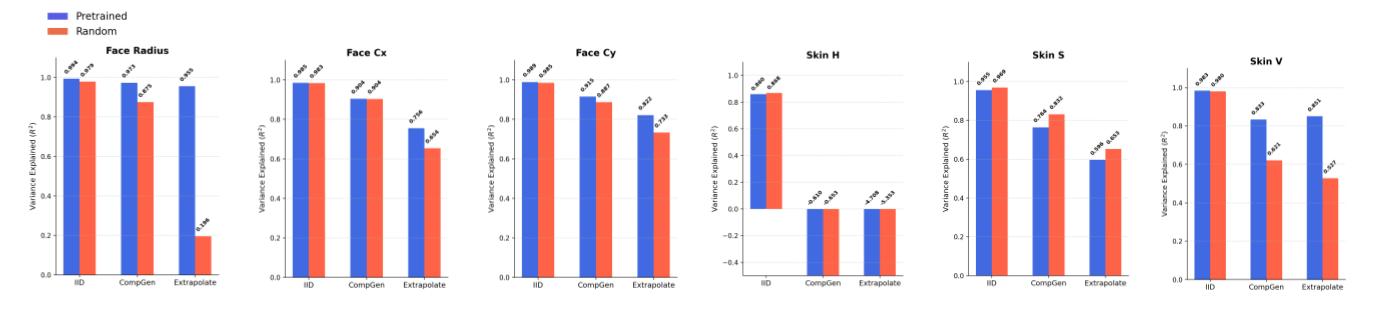}
\caption{\textbf{SVG-Faces: generalization within one style.} Per-latent test $R^2$ for six latents in three regimes: \textbf{IID} (random split), \textbf{CompGen} (train on one half of a latent's range, test on the other half) and \textbf{Extrapolate} (train within a sub-range, test outside it). The pretrained encoder extrapolates better on face radius and skin value; both encoders fail on skin hue outside the IID split, and random initialization is ahead on skin saturation.}
\label{fig:ood-faces}
\end{figure}

\paragraph{Generalization within one style.} With a single visual style, the natural out-of-distribution test restricts the range of one latent during training. Fig.~\ref{fig:ood-faces} reports six latents in three regimes: IID (random split), CompGen (train on one half of the latent's range, test on the other half) and Extrapolate (train within a sub-range, test outside it). On geometry the pretrained encoder degrades less than random initialization; the clearest case is face radius under extrapolation ($R^2 = 0.96$ against $0.20$). Skin value behaves the same way. Skin hue fails for both encoders outside the IID split, with negative $R^2$, and on skin saturation random initialization is ahead in every regime. Pretrained representations thus extrapolate better on geometry but not on color, in line with reports that pretrained representations often fail to generalize compositionally within a domain \citep{schott2022visual}. The cross-style test of \S\ref{sec:ood} replaces this range restriction on SVG-World.

% \clearpage

% NeurIPS 2026 reproducibility checklist (included at submission).
% \input{checklist.tex}

\end{document}